\documentclass[11pt]{article}

\usepackage{acl}

\newcommand{\std}[1]{_{\pm\scriptscriptstyle #1}}
\usepackage{float}
\usepackage{enumitem}
\usepackage{booktabs,tabularx}
\usepackage{multirow}

\usepackage{microtype}
\usepackage{graphicx}     % for \includegraphics
\usepackage{subcaption}   % for the subfigure environment and sub-
\usepackage[english,bidi=default]{babel} % English as 
\usepackage{amsmath,amssymb}

\usepackage{algpseudocode}
\usepackage{comment}
\usepackage{soul}
\sethlcolor{yellow}
\usepackage{capt-of}
\usepackage{booktabs}
\usepackage[ruled,vlined,linesnumbered,noend]{algorithm2e}
\SetKwInOut{Require}{Require}
\SetAlgoNlRelativeSize{-1}
\usepackage{color-edits}
\addauthor{aa}{blue}
\usepackage{varwidth}   
\usepackage{xcolor}

\usepackage[most]{tcolorbox}
\tcbuselibrary{listings,skins,breakable}
\usepackage{listings}

\newtcblisting{examplequery}[1]{%
  enhanced,
  breakable,
  title={#1},
  fonttitle=\bfseries,
  coltitle=white,
  colback=black!5,
  colbacktitle=black!80,
  colframe=black!60,
  arc=2mm,
  boxrule=0.4pt,
  left=6pt,right=6pt,top=6pt,bottom=6pt,
  listing engine=listings,
  listing only,
  listing options={style=promptmd},
  width=\textwidth,          % full width of the page
  float,                     % make it a float
  float*=htbp,               % STAR = span two columns (like figure*)
  floatplacement=htbp,       % placement hints
}

\lstdefinestyle{promptmd}{
  basicstyle=\ttfamily\footnotesize,
  breaklines=true, columns=fullflexible, keepspaces=true,
  moredelim=**[is][\color{red}]{<<}{>>} % mark spans to color
}

\title{LLM Prompt Duel Optimizer: Efficient Label-Free Prompt Optimization} 

\newif\ifarxiv
\arxivtrue

\author{
\textbf{Yuanchen Wu\textsuperscript{1}}\thanks{Work done during an internship at Meta.}\thanks{Code is available at \url{https://github.com/meta-llama/prompt-ops}}  \quad
\textbf{Saurabh Verma\textsuperscript{2}} \quad
\textbf{Justin Lee\textsuperscript{2}} \quad
\textbf{Fangzhou Xiong\textsuperscript{2}} \\
\textbf{Poppy Zhang\textsuperscript{2}} \quad
\textbf{Amel Awadelkarim\textsuperscript{2}} \quad
\textbf{Xu Chen\textsuperscript{2}} \quad
\textbf{Yubai Yuan\textsuperscript{1}} \quad
\textbf{Shawndra Hill\textsuperscript{2}}\\
\textsuperscript{1}Department of Statistics, The Pennsylvania State University\quad
\textsuperscript{2}Meta
}
\begin{document}

\maketitle
\begin{abstract}
Large language models (LLMs) are highly sensitive to prompts, but most automatic prompt optimization (APO)  methods assume access to ground-truth references (e.g., labeled validation data) that are costly to obtain. We propose the \textbf{Prompt Duel Optimizer (PDO)}, a sample-efficient framework for \emph{label-free} prompt optimization based on pairwise preference feedback from an LLM judge. PDO casts prompt selection as a dueling-bandit problem and combines (i) \emph{Double Thompson Sampling} to prioritize informative comparisons under a fixed judge budget, with (ii) \emph{top-performer guided mutation} to expand the candidate pool while pruning weak prompts. Experiments on BIG-bench Hard (BBH) and MS MARCO show that PDO consistently identifies stronger prompts than label-free baselines, while offering favorable quality--cost trade-offs under constrained comparison budgets.
\end{abstract}

\section{Introduction}
LLM performance hinges on well-crafted prompts that unlock their capabilities \citep{sun2023zeroshot}. Creating effective prompts typically requires extensive trial-and-error or task-specific techniques (e.g., chain-of-thought prompting for reasoning tasks; \citealp{wei2022chainofthought}), which often do not transfer across tasks or domains. This limitation motivates Automatic Prompt Optimization (APO) \citep{ramnath2025systematic}, which iteratively generates and evaluates candidate prompts to discover high-performing instructions. 

Despite encouraging results across diverse tasks, most APO methods \citep{zhou2022ape,yang2024opro,fernando2023promptbreeder,pryzant2023protegi} rely on \textbf{reference-based supervision} (e.g., ground-truth labels, gold references, or label-derived rewards) to score candidates on validation sets. In many practical settings, however, obtaining such supervision at scale is costly and slow \citep{ratner2017label}. Still, prompt optimization remains essential; for example, in industrial LLM-based text classification, practitioners need reasonably good prompts to initiate deployment before large human-labeled datasets are available \citep{wagner2024usingllms}. Such challenges raise a central question:

\begin{itemize}[label=\scalebox{1}{$\bullet$}]
    \item \textit{Can we optimize prompts without ground-truth label references?}
\end{itemize}

One direction is to reduce reliance on human annotation by adopting automatic evaluation of model outputs. Recent work suggests that LLMs can serve as judges of model outputs, including in reference-free settings \citep{zheng2023mtbench,liu-etal-2023-geval,gu2024llmjudgeSurvey}. As prompt quality is reflected in generated responses, evaluating prompts with an LLM judge is a natural approach. In this setting, rather than scoring each output independently, it is often preferable to rely on \textbf{pairwise preference feedback}: a judge compares outputs from two prompts on the same input and selects the preferred one. Pairwise comparisons typically yield a more reliable signal than direct pointwise scoring, which is prone to calibration issues \cite{liu2024pairwiseCOLM}.  

LLM preference feedback introduces two challenges. \emph{First}, LLM judges are noisy: calls are non-deterministic \citep{he2025nondeterminism}, judgments can exhibit position and verbosity biases \citep{zheng2023mtbench}, and task complexity may amplify these effects. \emph{Second}, preference-based evaluation scales quadratically with the number of candidates; each comparison requires an LLM API call and incurs monetary cost, making exhaustive evaluation infeasible.

To address these issues, we introduce the \textbf{Prompt Duel Optimizer (PDO)}, which casts label-free prompt optimization as a dueling-bandit problem where the cost is the number of judge comparisons. PDO combines \emph{Double Thompson Sampling (D-TS)} to focus comparisons on informative prompt pairs with a top-performer \emph{mutate-and-prune} strategy that expands the pool while discarding weak candidates. We further assess judge reliability via cross-family re-evaluation to probe circularity and show that PDO’s gains are robust across judge choices.

We summarize our contributions as follows:
\begin{enumerate}
  \item We formalize label-free prompt optimization with LLM-judged pairwise comparisons as a dueling-bandit problem.
  \item We propose PDO, integrating D-TS for sample-efficient prompt selection with top-performer guided mutation that steers search toward stronger regions and yields better candidates.
  \item We show empirically on BBH and MS MARCO that PDO outperforms label-free baselines under fixed judge budgets, and analyze quality--cost trade-offs via budget and runtime comparisons.
  \item We analyze judge noise and potential circularity, and demonstrate that PDO’s gains persist under alternative judge families.
\end{enumerate}

\section{Preliminaries and Problem Setup}

In this section, we establish the connection between preference-based prompt optimization and dueling bandits.
\subsection{Background: Dueling Bandits}
In the $K$-armed dueling bandit problem \citep{2021DB_Survey}, at each round the decision-maker selects two arms $i$ and $j$ to duel, and the comparison yields a stochastic preference outcome. Specifically, the probability that $i$ is preferred to $j$ is
\[
\mu(i,j) = \Pr(i \succ j).
\]
A \emph{Condorcet winner} is an arm $i^\ast$ such that $\mu(i^\ast,j) > 0.5$ for all $j \neq i^\ast$. When no Condorcet winner exists, a standard choice rule is the Copeland criterion: the \emph{Copeland score} of arm $i$ counts the number of opponents it beats with probability greater than $1/2$:
\[
C(i) = \bigl\lvert \{\, j \neq i : \mu(i,j) > \tfrac{1}{2} \,\} \bigr\rvert.
\]
A \emph{Copeland winner} is then an arm $i^\star \in \arg\max_{i} C(i)$.

% We study the problem of optimizing discrete prompts for a black-box large language model (LLM) in settings where no ground-truth task labels or scalar rewards are available. Instead, supervision comes from a \emph{preference judge}, which compares outputs of the LLM under two different prompts on the same input and indicates which one is preferred. This setting reflects practical scenarios such as alignment, subjective quality assessment, or preference elicitation, where scalar rewards are infeasible but pairwise feedback is natural.

\subsection{Prompt Optimization Through the Lens of Dueling Bandits}
Building on these definitions, we formalize preference-based prompt optimization in the dueling bandits setting.

\paragraph{Problem Setup.}  
Let $\mathcal{X}$ denote the input space and $\mathcal{P}=\{p_1,\ldots,p_K\}$ a finite set of candidate prompts. Let $D_{\mathcal{X}}$ be a distribution over $\mathcal{X}$. For $p \in \mathcal{P}$ and $x \sim D_{\mathcal{X}}$, let $f_p(x)$ denote the LLM output when applying prompt $p$ to input $x$. To compare the two prompts $p_i,p_j\in\mathcal{P}$ on the same input $x$, we query an LLM judge and record a binary preference:
\[
\mathrm{Judge}_x(p_i,p_j) =
\begin{cases}
1, & \text{if } f_{p_i}(x) \succ f_{p_j}(x),\\
0, & \text{otherwise.}
\end{cases}
\]
Given a set of unlabeled examples $\{x_i\}_{i=1}^n$, the empirical estimate of $\mu(p_i,p_j)$ becomes
\[
\widehat{\mu}(p_i,p_j) 
\;=\; \frac{1}{n}\sum_{k=1}^n \mathbf{1}\!\big[\mathrm{Judge}_{x_k}(p_i,p_j)=1\big].
\]

\paragraph{Prompt Optimization Objective.}
The goal of prompt optimization is to identify a prompt that maximizes task performance. In the absence of ground-truth references, we use pairwise preferences as a practical proxy for selecting high-quality prompts. Using empirical estimates $\widehat{\mu}(p_i,p_j)$, we select the Condorcet winner when it exists, or otherwise the Copeland winner. In practice, limited API budgets make exhaustive estimation ($O(n|\mathcal{P}|^2)$ comparisons) infeasible, motivating sample-efficient methods that adaptively target informative pairs while aligning with the Condorcet/Copeland objectives.

% This formulation connects the theoretical notion of optimality with the practical need for sample efficiency in preference-based prompt optimization.

\begin{figure}
    \centering
    \includegraphics[width=\linewidth]{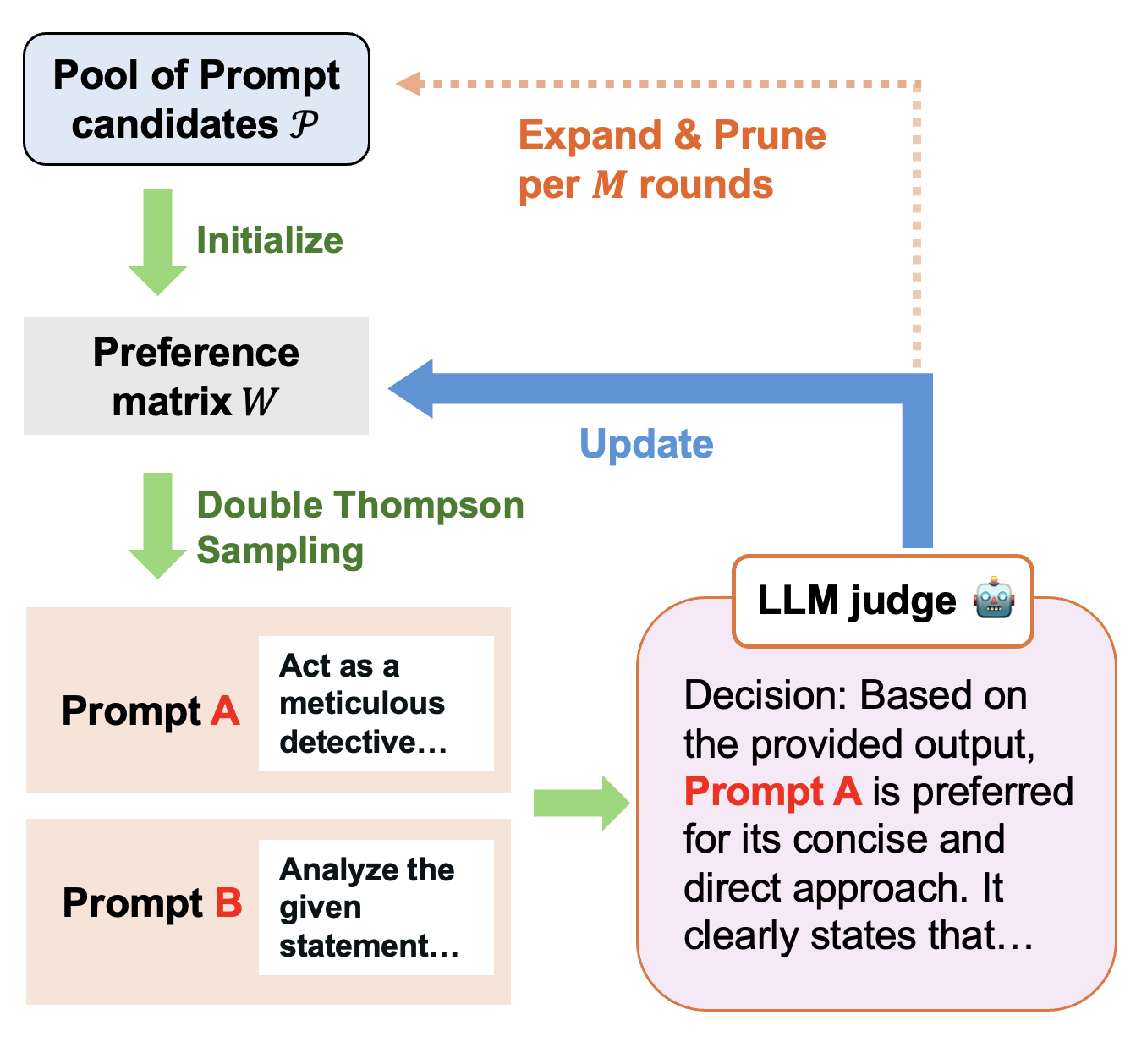}
    \caption{Workflow of the Prompt Duel Optimizer.}
    \label{fig:enter-label}
\end{figure}

\section{Prompt Duel Optimizer (PDO)}
To address the above challenge, we propose the \textbf{Prompt Duel Optimizer (PDO)}, an algorithm designed to identify high-performing prompts under limited comparison budgets. PDO combines two complementary components: (1) \emph{Double Thompson Sampling (D-TS)} \citep{wu2016double}, a Bayesian strategy for efficient pairwise evaluation that targets Copeland-optimal winners; and (2) \emph{Top-Performer Guided Mutation}, which adaptively mutates top-performing prompts to expand coverage of the search space.

\subsection{Efficient Prompt Selection via Double Thompson Sampling}

D-TS extends Thompson Sampling \citep{agrawal2012ts} to the dueling bandit setting, where feedback comes from pairwise comparisons rather than scalar rewards. Each round uses two independent Thompson draws to concentrate queries on informative duels and guides the search toward a \emph{Copeland-optimal} prompt.

\paragraph{Notations.}
For each pair $(p_i,p_j)$, let $W_{ij}$ denote the current number of wins of $p_i$ over $p_j$, and let $N_{ij} = W_{ij}+W_{ji}$ be the total number of duels. 
The Bayesian posterior for the probability that $p_i$ beats $p_j$ is modeled as $\theta_{ij} \sim \mathrm{Beta}(W_{ij}+1,\, W_{ji}+1)$. For $\alpha > 0$, the corresponding upper and lower confidence bounds are defined as
\[
\begin{aligned}
u_{ij}(t) &= \frac{W_{ij}}{N_{ij}} + \sqrt{\frac{\alpha \log t}{\max\{1, N_{ij}\}}}, \\
l_{ij}(t) &= \frac{W_{ij}}{N_{ij}} - \sqrt{\frac{\alpha \log t}{\max\{1, N_{ij}\}}}.
\end{aligned}
\]

\paragraph{Per-round Procedure ($t=1,2,\dots,T$):}
\begin{enumerate}
  \setlength\itemsep{0pt}
  \setlength\parskip{0pt}
  \setlength\parsep{0pt}
\item \emph{First Prompt Selection.}  
Compute an optimistic Copeland score for each prompt:
\[
\widehat \zeta_i(t)=\tfrac{1}{K-1}\sum_{j\neq i}\mathbf{1}\{u_{ij}(t)\ge 0.5\}
\]
Keep only prompts with the maximum score in a set $\mathcal{\zeta}(t) = \{\, i \;|\; \widehat \zeta_i(t) = \max_{k} \widehat \zeta_k(t) \,\}$.  
For each $i\in \mathcal{\zeta}(t)$, draw independent samples $\theta^{(1)}_{ij}$ and count
\[
s_i=\sum_{j\neq i}\mathbf{1}\{\theta^{(1)}_{ij}\ge 0.5\}
\]
Select prompt $i^\star=\arg\max_{i\in \mathcal{\zeta}(t)} s_i$.

\item \emph{Second Prompt Selection.}  
Restrict to only uncertain opponents for $i^\star$,
\[
S_{i^\star}(t)=\{j\neq i^\star:\; l_{i^\star j}(t)\le 0.5 \}
\]
Draw independent samples $\theta^{(2)}_{j i^\star}$ and select $j^\star=\arg\max_{j\in S_{i^\star}(t)} \theta^{(2)}_{ji^\star}$

\item \emph{Duel and update.}  
Judge prompts $(p_{i^\star},p_{j^\star})$, record the winner, and update $W_{i^\star j^\star}$.
\end{enumerate}

% This two-step prompt selection procedure using D-TS achieves the following regret bounds in general Copeland settings.

% \begin{theorem}[PDO Copeland Regret]
% \label{thm:dts-regret}
% Let $\mathcal{P}=\{p_1,\dots,p_K\}$ be $K$ prompts with pairwise preference probabilities $\mu_{i,j}=\Pr(p_i\succ p_j)$. 
% Define the normalized Copeland score
% \[
% \zeta_i=\tfrac{1}{K-1}\sum_{j\neq i}\mathbf{1}\{\mu_{i,j}>1/2\},\quad 
% \zeta^*=\max_i \zeta_i.
% \]
% If at round $t$ the algorithm selects $(a_t,b_t)$, the regret is 
% $r_t=\zeta^*-\max\{\zeta_{a_t},\zeta_{b_t}\}$ and $R_T=\sum_{t=1}^T r_t$.  
% Then Double Thompson Sampling (D-TS) achieves \citep{wu2016double}
% \[
% \mathbb{E}[R_T]=O(K^2\log T).
% \]
% \end{theorem}

% \paragraph{Implications.}
% The theorem implies that $\tfrac{\mathbb{E}[R_T]}{T}\!\to\!0$ as $T\!\to\!\infty$, i.e., PDO with D-TS asymptotically converges to selecting Copeland-optimal prompts. 
% If a Condorcet winner exists, D-TS converges to that unique prompt; otherwise, it converges to the set of Copeland winners.

\subsection{Efficient Prompt Discovery via Top-Performer Guided Mutation}

While D-TS identifies the Copeland-optimal prompt within a fixed pool, the broader goal of PDO is to locate the global optimum $p^*$ over a combinatorially large prompt space. Because exhaustive search is infeasible, PDO, inspired by evolutionary prompt optimization techniques \citep{fernando2023promptbreeder,guo2023evoprompt}, incrementally expands the candidate pool by mutating top-performing prompts and introducing mutation tips that promote variation and diversity without straying too far from high-performing prompts.

\paragraph{Mutation Procedure.} At round $t$, let the current pool be 
$\mathcal{P}_t = \{p_1, \dots, p_{K_t}\}$ with the empirical Copeland scores $\widehat{\mathbf{C}}_t:=\bigl(\widehat{C}_t(p)\bigr)_{p\in\mathcal{P}_t}$. The procedure is:
\begin{enumerate} 
  \setlength\itemsep{0pt}
  \setlength\parskip{0pt}
  \setlength\parsep{0pt}
    \item \textit{Selection:} Choose the top-performing prompt $p^*_t = \arg\max_{p \in \mathcal{P}_t} \widehat{\mathbf{C}}_t$.
    \item \textit{Mutation:} Generate a variant $p_{\text{new}}$ of $p^*_t$ via template edits, text-gradient guided changes, or LLM-assisted rewrites.
    \item \textit{Expansion:} Update pool  $\mathcal{P}_{t+1} = \mathcal{P}_t \cup \{p_{\text{new}}\}$.
    % \item \textbf{Evaluation:} estimate $\hat{f}(p')$ through pairwise comparisons using D-TS.
\end{enumerate}

\subsection{Theoretical insights of PDO}
PDO can be viewed as a two-level optimization scheme that alternates \emph{selection} and \emph{discovery}.
Within any fixed candidate pool, D-TS provides a principled way to allocate pairwise comparisons and is known to achieve $O(K^2\log T)$ expected Copeland regret in general settings, implying vanishing average regret and convergence to the Copeland-winner set as the comparison budget grows \citep{wu2016double}.

To move beyond a fixed pool, PDO periodically expands the candidate set by mutating the current Copeland leader.
Under a standard smoothness assumption that prompt utility varies gradually under local edits (e.g., $L$-Lipschitz with respect to a prompt distance), this strategy concentrates new candidates in increasingly competitive neighborhoods, analogous to the ``zooming-in'' intuition in Lipschitz/metric bandits \citep{KleinbergSlivkinsUpfal2008}.

Taken together, PDO separates prompt optimization into (i) sample-efficient identification of the best prompt among current candidates (via D-TS) and (ii) targeted exploration that increases coverage of near-optimal regions (via top-performer mutation), which helps explain its empirical efficiency under limited judge budgets.

\algrenewcommand\algorithmiccomment[1]{\hfill{\footnotesize$\triangleright$~#1}}
\makeatletter
% fixed-width line-number gutter so wrapped lines don't jiggle numbers
\algrenewcommand\alglinenumber[1]{\scriptsize\makebox[1.8em][l]{#1:}}
% slightly smaller left indent so long lines don't wrap early
\makeatother

\begin{algorithm}[t]
\caption{Prompt Duel Optimizer}
\label{alg:pdo_simple2}
\Require{$\mathcal{P}_0$ (initial pool); judge $\mathcal{J}$; batch size $m$; total rounds $T$; mutation period $M$}
$\mathcal{P} \leftarrow \mathcal{P}_0$; initialize $W,N \in \mathbb{R}^{|\mathcal{P}|\times|\mathcal{P}|}$ to zeros\;
\For{$t \leftarrow 1$ \KwTo $T$}{
  $(p_i,p_j) \leftarrow \textsc{D-TS}(\mathcal{P},W,N)$ \tcp*{prompt selection via D-TS}
  $(w_i,w_j) \leftarrow \textsc{Duel}(p_i,p_j,m,\mathcal{J})$ \tcp*{get win counts}
  $W[i,j] \leftarrow W[i,j]+w_i$\; $W[j,i] \leftarrow W[j,i]+w_j$\;
  $N[i,j] \leftarrow N[i,j]+m$\;  $N[j,i] \leftarrow N[j,i]+m$\;
  \If{$t \bmod M = 0$}{
     $p_{t}^\star \leftarrow \textsc{CurrentBest}(\mathcal{P},W,N)$ \tcp*{Copeland rank}
     $p_{\text{new}} \leftarrow \textsc{Mutate}(p_{t}^\star)$ \tcp*{generate new prompts via top-performer guided mutation}
     $\mathcal{P} \leftarrow \mathcal{P}\cup\{p_{\text{new}}\}$\; Expand $W,N$ with zero row/column for $p_{\text{new}}$\;
  }
}
\Return $\textsc{CurrentBest}(\mathcal{P},W,N)$ \tcp*{final Copeland winner}
\end{algorithm}

\begin{table*}[htbp]
\centering
\scriptsize
\setlength{\tabcolsep}{3pt}%
\renewcommand{\arraystretch}{1.2}
\begin{tabular}{l|cccccccc}
\hline
\textbf{Method} & \textbf{Causal} & \textbf{Date} & \textbf{DisambigQA} & \textbf{Formal} & \textbf{Geometric} & \textbf{Hyperbaton} & \textbf{Logical-5} & \textbf{Logical-7} \\
\hline
No prompt & \( \underline{0.661} \std{0.044} \) & \( 0.854 \std{0.024} \) & \( 0.698 \std{0.047} \) & \( \underline{0.739} \std{0.031} \) & \( 0.434 \std{0.036} \) & \( \underline{0.900} \std{0.020} \) & \( 0.785 \std{0.018} \) & \( \underline{0.739} \std{0.033} \) \\
CoT       & \( 0.653 \std{0.042} \) & \( 0.877 \std{0.019} \) & \( 0.720 \std{0.039} \) & \( 0.725 \std{0.027} \) & \( 0.422 \std{0.028} \) & \( 0.891 \std{0.023} \) & \( 0.761 \std{0.027} \) & \( 0.726 \std{0.025} \) \\
PoS       & \( 0.652 \std{0.037} \) & \( 0.878 \std{0.019} \) & \( 0.698 \std{0.043} \) & \( \underline{0.739} \std{0.027} \) & \( 0.403 \std{0.030} \) & \( 0.896 \std{0.024} \) & \( \underline{0.798} \std{0.032} \) & \( \mathbf{0.750} \std{0.026} \) \\
SPO       & \( 0.655 \std{0.033} \) & \( \underline{0.884} \std{0.017} \) & \( \underline{0.725} \std{0.057} \) & \( 0.738 \std{0.018} \) & \( \mathbf{0.650} \std{0.069} \) & \( 0.886 \std{0.031} \) & \( 0.787 \std{0.031} \) & \( 0.721 \std{0.026} \) \\
PDO (ours)      & \( \mathbf{0.681} \std{0.040} \) & \( \mathbf{0.918} \std{0.014} \) & \( \mathbf{0.738} \std{0.050} \) & \( \mathbf{0.744} \std{0.030} \) & \( \underline{0.598} \std{0.073} \) & \( \mathbf{0.910} \std{0.029} \) & \( \mathbf{0.804} \std{0.034} \) & \( 0.711 \std{0.019} \) \\
\hline
\textbf{Method} & \textbf{Navigate} & \textbf{Penguins} & \textbf{Salient} & \textbf{Snarks} & \textbf{Tracking-5} & \textbf{Tracking-7} & \textbf{Tracking-3} & \textbf{Web of Lies} \\
\hline
No prompt & \( 0.869 \std{0.013} \) & \( 0.915 \std{0.018} \) & \( \underline{0.698} \std{0.020} \) & \( 0.823 \std{0.027} \) & \( 0.695 \std{0.033} \) & \( 0.499 \std{0.020} \) & \( 0.890 \std{0.019} \) & \( 0.766 \std{0.020} \) \\
CoT       & \( \underline{0.878} \std{0.016} \) & \( 0.915 \std{0.023} \) & \( \mathbf{0.709} \std{0.021} \) & \( \underline{0.833} \std{0.023} \) & \( 0.724 \std{0.046} \) & \( 0.532 \std{0.025} \) & \( \underline{0.904} \std{0.019} \) & \( 0.796 \std{0.022} \) \\
PoS       & \( 0.866 \std{0.019} \) & \( 0.910 \std{0.027} \) & \( 0.693 \std{0.025} \) & \( 0.816 \std{0.026} \) & \( \underline{0.725} \std{0.030} \) & \( 0.538 \std{0.034} \) & \( 0.888 \std{0.019} \) & \( \underline{0.861} \std{0.019} \) \\
SPO       & \( 0.874 \std{0.035} \) & \( \underline{0.934} \std{0.025} \) & \( 0.662 \std{0.038} \) & \( 0.820 \std{0.046} \) & \( 0.692 \std{0.046} \) & \( \underline{0.543} \std{0.026} \) & \( 0.826 \std{0.087} \) & \( 0.818 \std{0.043} \) \\
PDO (ours)      & \( \mathbf{0.900} \std{0.023} \) & \( \mathbf{0.937} \std{0.034} \) & \( 0.681 \std{0.032} \) & \( \mathbf{0.840} \std{0.039} \) & \( \mathbf{0.796} \std{0.084} \) & \( \mathbf{0.641} \std{0.089} \) & \( \mathbf{0.930} \std{0.046} \) & \( \mathbf{0.942} \std{0.040} \) \\
\hline
\end{tabular}
\caption{Test results on 16 BBH tasks averaged over 10 runs. For PDO, we report the test accuracy of the prompt selected by the highest Copeland score. PDO is compared with baselines that assume \textbf{no access} to ground-truth labels. The best performance is shown in bold, and the second-best is underlined.}
\label{Without_label}
\end{table*}

\section{Experiment and Results}
\subsection{Experimental Setup}

\paragraph{Datasets.} In this section, we conduct experiments to evaluate PDO on both closed-ended multiple-choice tasks and open-ended QA tasks. For the multiple-choice setting, we select 16 tasks from Big-Bench-Hard \citep{suzgun2022BBH}, using accuracy as the evaluation metric. For the open-ended QA setting, we consider four task categories from MS-MARCO \citep{bajaj2016msmarco}, where the final evaluation metric is an integer score between $1$ and $5$, assigned by an LLM judge comparing the model’s output with the ground-truth answer provided by the original dataset.  

Across both task types, for the results reported in this section, we randomly split the data into development and test sets with a 50/50 split ratio. We report the test set performance averaged over 10 runs. The \textit{Llama-3.3-70B-Instruct} model is used for prompt generation, preference judging, and final evaluation. Detailed descriptions of the datasets and the experimental setup of PDO are provided in Appendix \ref{Experiment_details}.

\paragraph{LLM Judge Design.} For multiple-choice tasks, we require the LLM to produce both an answer and the accompanying reasoning given the instruction prompt. We then apply a dual-judge approach: if two prompts yield different answers, the LLM judge selects the prompt with the correct answer; if they yield the same answer, the judge decides based on the quality of reasoning. For open-ended tasks, the LLM judge design is more straightforward: each pair of responses is evaluated on accuracy, completeness, relevance, and clarity, consistent with the criteria of the final evaluation metric. To mitigate position bias, the order of the two outputs in each prompt pair is randomized before being fed into the judge template. A detailed rationale and analysis of the LLM preference judge design are provided in Section~\ref{judge} and Appendix~\ref{appendix_judge_design}. 

\begin{figure*}[t]
\centering
\begin{subfigure}[b]{0.24\textwidth}
    \centering
    \includegraphics[width=\textwidth]{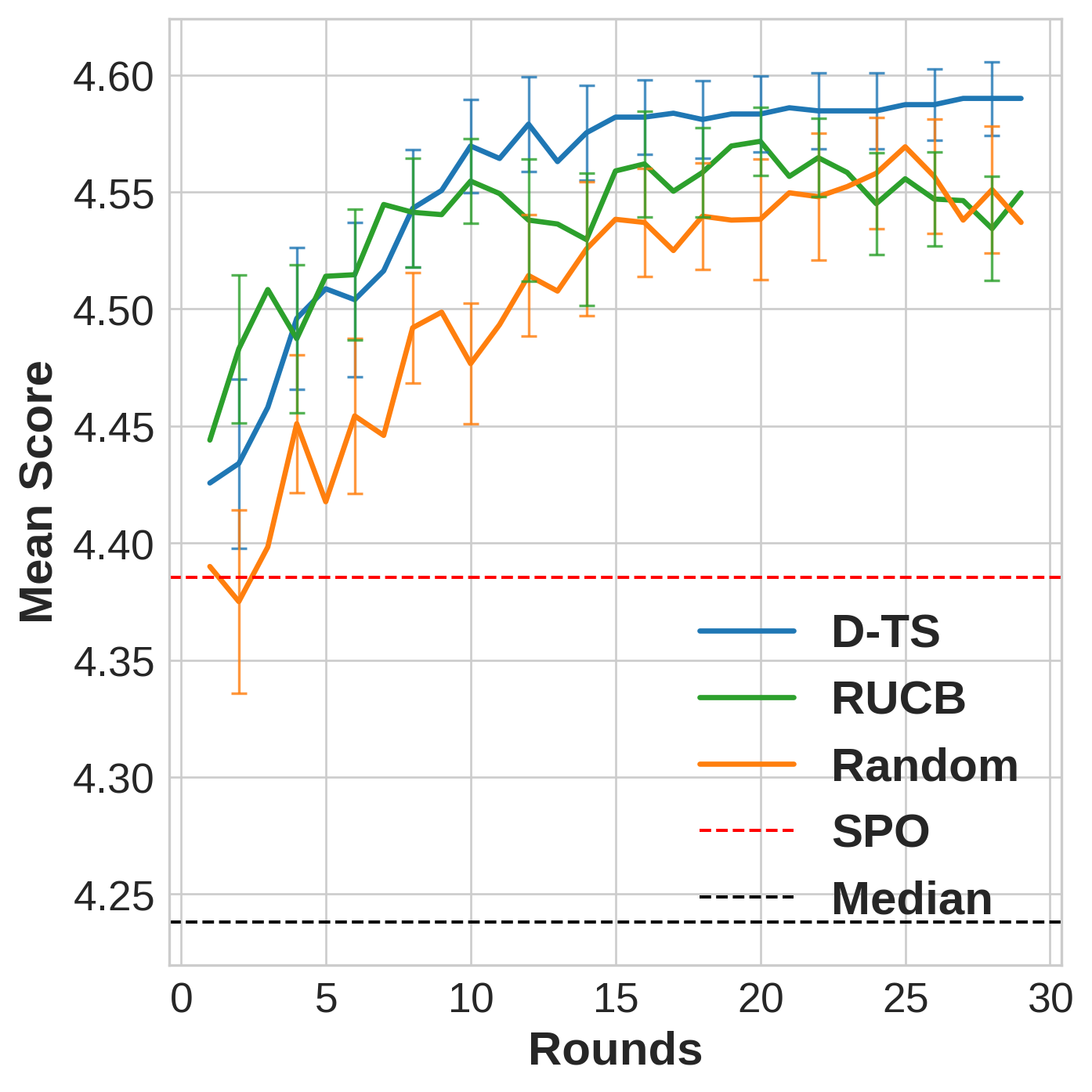}
    \caption{Description}
    \label{fig:description}
\end{subfigure}\hfill
\begin{subfigure}[b]{0.24\textwidth}
    \centering
    \includegraphics[width=\textwidth]{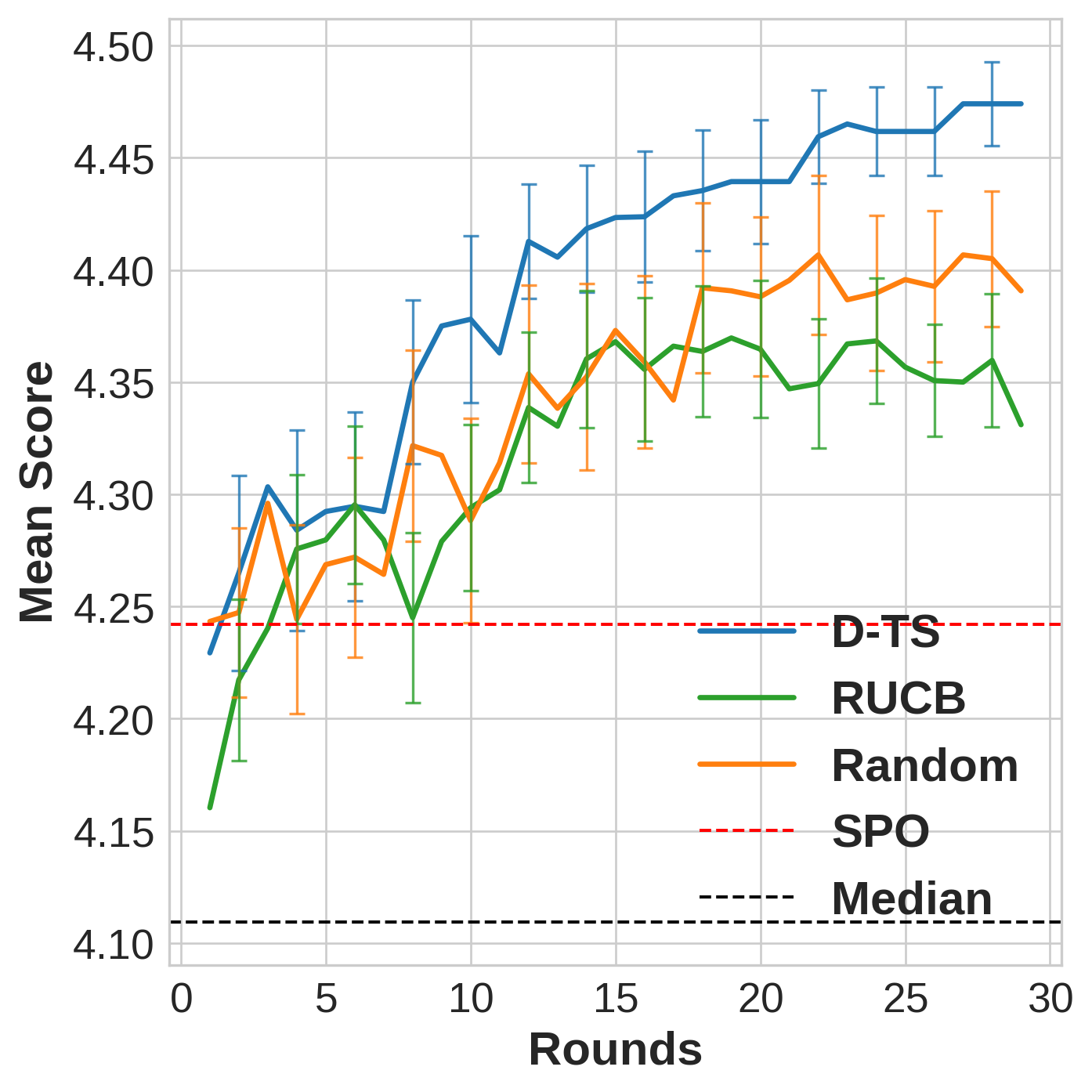}
    \caption{Entity}
    \label{fig:entity}
\end{subfigure}\hfill
\begin{subfigure}[b]{0.24\textwidth}
    \centering
    \includegraphics[width=\textwidth]{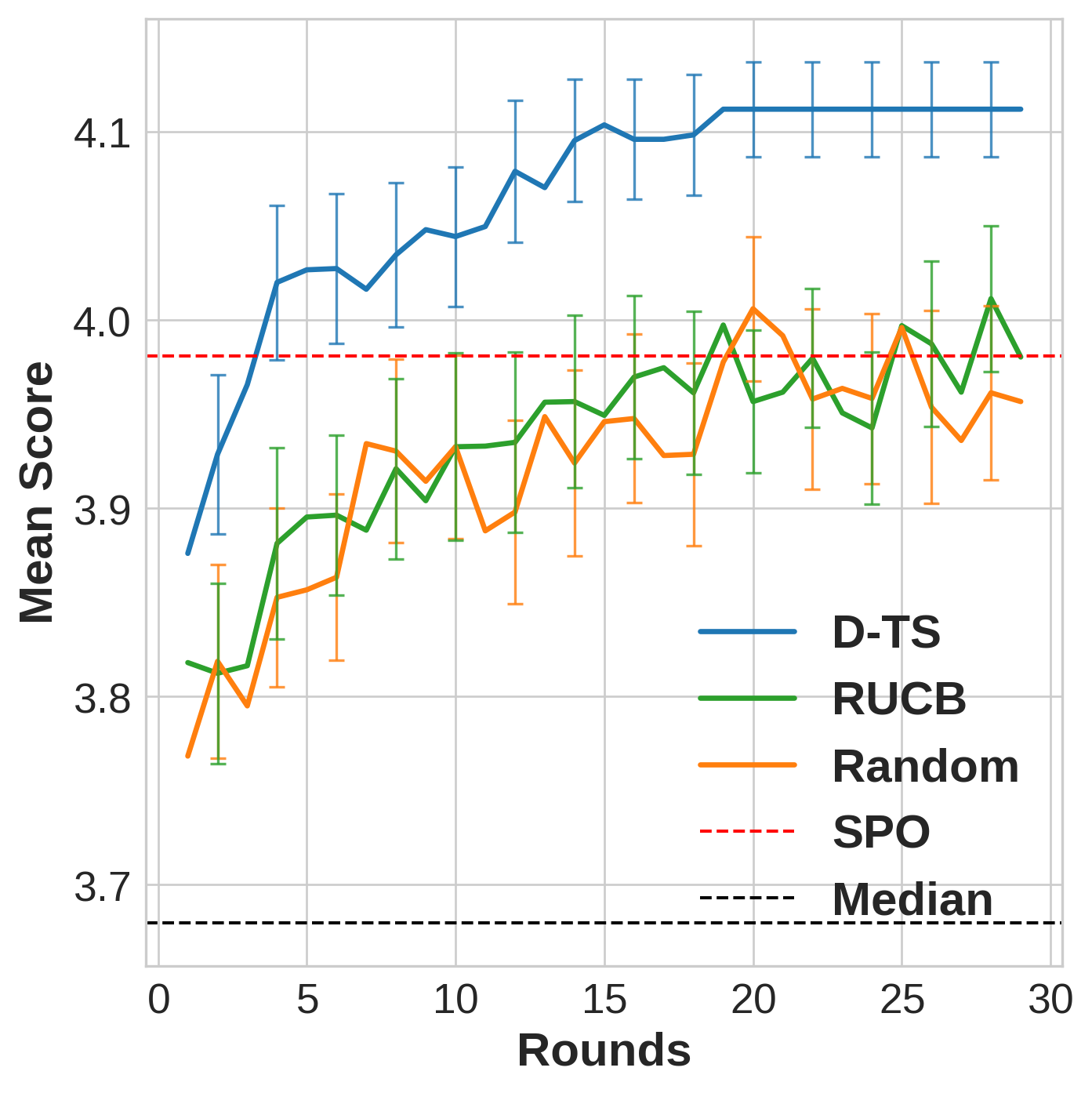}
    \caption{Numeric}
    \label{fig:numeric}
\end{subfigure}\hfill
\begin{subfigure}[b]{0.24\textwidth}
    \centering
    \includegraphics[width=\textwidth]{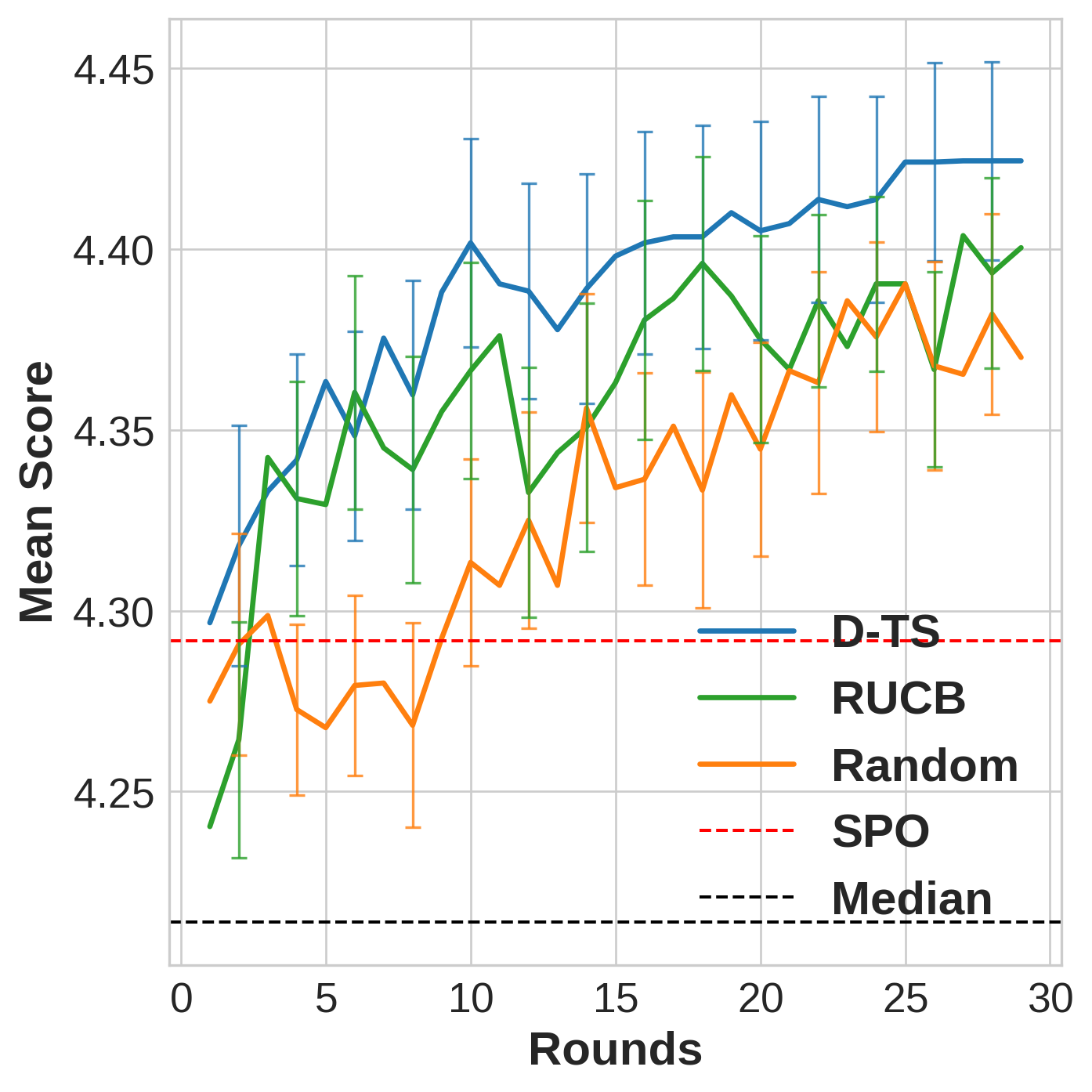}
    \caption{Location}
    \label{fig:location}
\end{subfigure}
\caption{Test performance of the winning prompt on the four MS-MARCO tasks. Each curve shows the mean score of the current Copeland leader over rounds, with 50 duels per round. PDO with D-TS consistently outperforms RUCB and Random sampling, and surpasses the SPO baseline within a few rounds. Gray lines indicate the median test score across all prompts generated by PDO.}
\label{fig:open_ended}
\end{figure*}

\paragraph{Baselines.} PDO is designed for prompt optimization without access to ground-truth labels or external references. A directly comparable baseline is \textbf{SPO}~\citep{Xiang2025SPO}, which similarly uses an LLM judge to iteratively compare the outputs of two prompts and select the winning prompt. We also include classical prompting techniques including chain-of-thought \textbf{COT} \citep{wei2022chainofthought} and plan-and-solve \textbf{PoS} \citep{wang-etal-2023-POS}.

For the Big-Bench-Hard (BBH) dataset, we also evaluate PDO under a supervised setting to enable comparison with popular supervised APO methods, including \textbf{APE}~\citep{zhou2022ape}, \textbf{OPRO}~\citep{yang2024opro}, and \textbf{Breeder}~\citep{fernando2023promptbreeder}. The optimization itself remains label-free; labels are introduced only in the final stage to select the prompt that achieves the highest development-set accuracy among candidates generated through initial proposals and prompt mutations. Results for the without-label and with-label settings are reported in Tables \ref{Without_label} and \ref{With_label}, respectively.

% Further details, including the full meta-prompt template, are provided in the Appendix.
% Morover, we further study the impact of introducing partial labels for PDO optimization process in section~\ref{With_label}.

\subsection{Benchmark Results}\label{main_result}

\paragraph{Multiple-choice Tasks Performance.}  As shown in Table \ref{Without_label}, using only preference signals from the LLM judge and selecting the winner prompt by Copeland scores, PDO achieves the highest evaluation accuracy on \textbf{13/16} tasks. Notable gains include \textit{Tracking-7} (0.641 vs. 0.543, \textbf{+9.8pp}) and \textit{Web of Lies} (0.942 vs. 0.861, \textbf{+8.1pp}).
% We examine the benefit of top-performer guided mutation in Appendix \ref{Prompt_mutation}.

% In the labeled setting, when the final prompt is chosen by development-set accuracy, PDO remains competitive with state-of-the-art prompt optimization baselines, achieving the top score on \textbf{9/16} tasks. For the \textit{Geometric} dataset, selecting the prompt based on development-set accuracy yields 0.712, outperforming all other methods, whereas selecting based on the LLM judge’s ranking results is only 0.598 accuracy, underperforming the label-free baseline SPO. In this particular case, this gap suggests that the LLM judge is less effective and noisier in identifying the best prompt on this  dataset compared with others in BBH. We further examine the issue of LLM judge noise in Section~\ref{judge}. 

\paragraph{Open-ended QA Tasks.}  
We evaluate four MS-MARCO tasks (\textit{Description}, \textit{Entity}, \textit{Numeric}, and \textit{Location}), starting from a pool of $|\mathcal{P}| = 50$ instructions. At each round $t$, we snapshot the win matrix $W_t$, compute Copeland scores, and report the test performance of the current Copeland winner. Figure~\ref{fig:open_ended} reports averages over 30 independent runs, comparing D-TS with the dueling-bandit alternative Relative Upper Confidence Bound (RUCB; \citealt{RUCB2014}), uniform Random sampling, and the SPO baseline. Across all tasks, D-TS achieves the highest scores and converges more quickly than RUCB and Random. The horizontal reference lines mark the median score of all prompts generated by PDO (gray) and the SPO baseline (red). Within just a few rounds, D-TS surpasses both reference lines and maintains this advantage over RUCB and Random. These results highlight the \emph{sample efficiency} of D-TS: it consistently identifies stronger prompts faster and more reliably than random sampling or alternative dueling-bandit methods.

\section{LLM Judge Analysis}\label{judge}

In the experiments reported above, we use two types of LLM judges. The first is the \emph{preference judge} used during LLM prompt optimization to compare outputs from two candidate prompts. The second is the \emph{evaluation judge}, which provides the final score for the open-ended MS-MARCO task. In this section, we present additional experiments to analyze the noise and potential bias introduced by both judge types.

% A fundamental bottleneck in PDO is that it depends on an LLM judge to approximate downstream performance through pairwise comparisons. In this section, we examine the factors that influence the effectiveness of the LLM judge in PDO.

\subsection{Preference Judge}\label{preference_judge}
\paragraph{Correlation Between Judge Noise and Task Performance.} We proxy preference judge effectiveness for the 16 BBH tasks by the test accuracy gap between Table~\ref{Without_label} (prompt chosen by Copeland score, label-free) and Table~\ref{With_label} (prompt chosen by development-set accuracy). In our results, \textit{Tracking-7} and \textit{Web of Lies} show small gaps and achieve top performance under both selection criteria, whereas \textit{Geometric} exhibits a large discrepancy of \textbf{11.4} percentage points. 

\begin{figure}[htbp]
    \centering
    % First subfigure
    \begin{subfigure}[b]{0.48\linewidth}
        \centering
        \includegraphics[width=\linewidth]{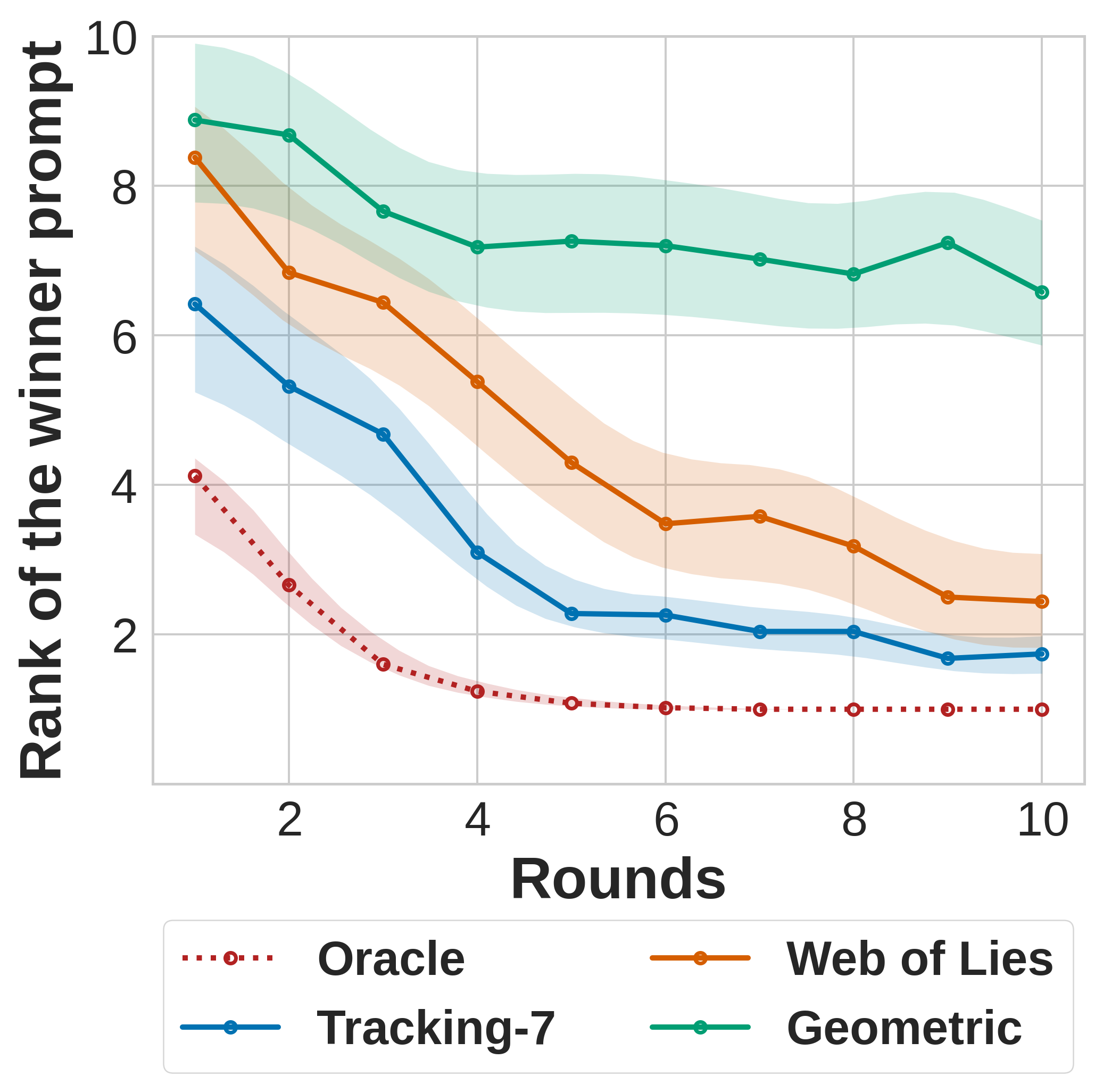}
        \caption{Judge noise across different BBH tasks.}
        \label{oracle_rank}
    \end{subfigure}
    \hfill
    % Second subfigure
    \begin{subfigure}[b]{0.48\linewidth}
        \centering
        \includegraphics[width=\linewidth]{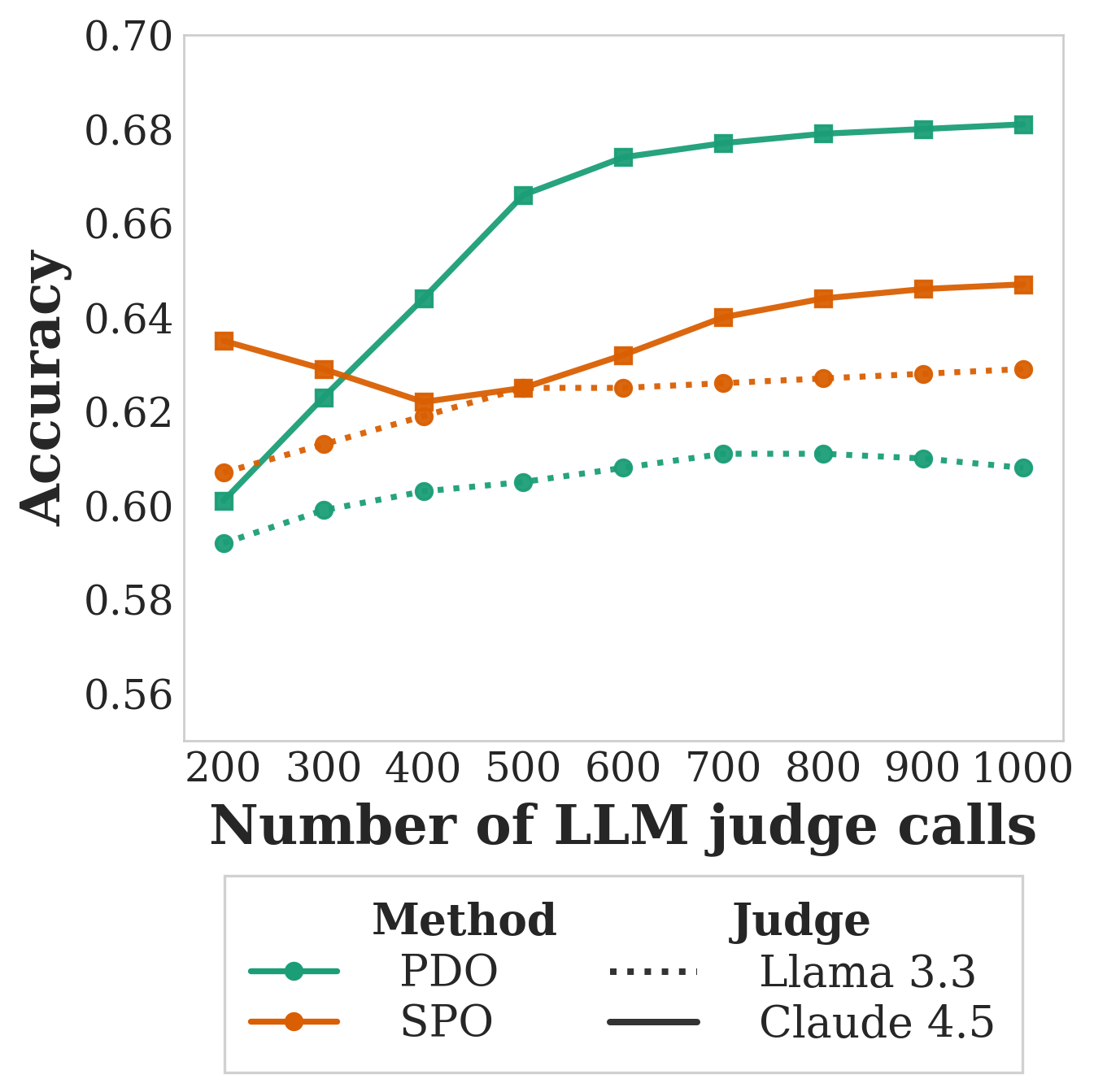}
        \caption{Noise reduction on \textit{Geometric} using Claude 4.5.}
        \label{claude_impact}
    \end{subfigure}
    
    \caption{(a) The accuracy of the LLM judge varies across tasks, with some tasks (e.g., \textit{Geometric}) showing persistent misjudgments compared to an oracle judge. (b) Using a frontier LLM judge (Claude 4.5) helps reduce the impact of noisy judgements.}
    \label{fig:oracle}
\end{figure}

To investigate further, for each task we fix $|\mathcal{P}|=20$ instruction prompts with varying accuracy, run D-TS with the LLM judge, and compare it against an \emph{oracle} judge that always selects the higher-accuracy prompt in each duel. At round $t$, we record the ground-truth accuracy rank of the current Copeland leader. Figure~\ref{oracle_rank} shows that the oracle converges to the best prompt by round 4; with the LLM judge, \textit{Tracking-7} steadily improves to rank 2 and \textit{Web of Lies} approaches rank $\approx 2.5$, while \textit{Geometric} remains around ranks 6–8 across rounds. These trends confirm that judge reliability is closely related to the performance gaps observed in Tables~\ref{Without_label} and~\ref{With_label}.

\paragraph{Reducing Judge Noise using frontier model.} 
We hypothesize that a key source of judge noise on BBH tasks arises from heterogeneous task difficulty: when two prompts produce different answers, the preference judge must determine which output is correct, and this decision becomes unreliable on harder tasks for judges with limited reasoning ability. A natural remedy is to replace the current Llama-3.3 preference judge with a stronger frontier model that better approximates an oracle, therefore reducing decision noise.

Table~\ref{tab:judge_model_results} reports the selection accuracy of the preference judge on examples where the two candidate answers differ. With Llama-3.3-70B as the judge, accuracy is high on \textit{Tracking-7} and \textit{Web of Lies} but only marginally above chance on \textit{Geometric}, consistent with the trend in Figure~\ref{oracle_rank}. Substituting Claude-4.5-Sonnet as the preference judge substantially improves selection accuracy on \textit{Geometric} and yields near-oracle performance on the other two tasks. As a result, when replacing the preference judge with Claude 4.5, the test accuracy of the prompts on \textit{Geometric} found by PDO exceeds that of SPO across LLM judge API budgets as reported in Figure \ref{claude_impact}. We report additional results on three other BBH tasks in Figure~\ref{fig:claude4.5more}.

\begin{table}[htbp]
\centering
\scriptsize % shrink font just for table
\setlength{\tabcolsep}{1.75pt} % very narrow spacing
\renewcommand{\arraystretch}{1.3}
\begin{tabular}{lccc}
\hline
\textbf{Judge Model} & \textbf{Geometric} & \textbf{Tracking-7} & \textbf{Web of Lies} \\
\hline
Llama-3.3-70B       & 0.59 & 0.89 & 0.85 \\
Claude-4.5-Sonnet   & 0.87 & 0.98 & 0.99 \\
\hline
\end{tabular}
\caption{Preference-judge selection accuracy on BBH instances where two candidate prompts produce different answers (1 = always selects the correct answer; 0.5 = random guessing).}
\label{tab:judge_model_results}
\end{table}

\subsection{Evaluation Judge}

\paragraph{Cross-family Evaluation.}
For the MS MARCO results in Section~4, we use the same Llama~3.3 model both to guide prompt optimization and to score the final prompt, which may introduce judge circularity (i.e., the optimized prompt is favored by the same model used for evaluation). To probe this issue beyond a single prompt pair, we form two prompt sets by sampling uniformly at random: 10 prompts produced by PDO and 10 prompts produced by SPO (all with their original Llama~3.3 judge scores recorded during optimization). To avoid additional inference variability, we use cached model outputs for each prompt on the same evaluation set and re-score these fixed outputs against the reference answers using judges from different model families.

Table~\ref{tab:judge_families} reports cross-family re-scoring results across the 10 sampled prompts per method. PDO remains higher than SPO under every judge, with consistently positive margins, suggesting that the observed improvement is not merely an artifact of circularity with the original Llama~3.3 judge.

In Appendix \ref{human_agreement}, we include a small-scale human agreement study testing whether the LLM judge evaluation on MS-MARCO aligns with human preferences on examples where PDO and SPO outputs receive different scores.

\begin{table}[htbp]
\centering
\scriptsize
\renewcommand{\arraystretch}{1.2}
\begin{tabular}{l|cc|c}
\hline
\textbf{Judge Model} & \textbf{PDO} & \textbf{SPO} & $\Delta$ \\
\hline
Llama 3.3 (original) & $4.68 \pm 0.06$ & $4.50 \pm 0.07$ & $0.18 $ \\
Mistral-Large        & $4.61 \pm 0.07$ & $4.49 \pm 0.07$ & $0.12 $ \\
GPT-4o               & $4.60 \pm 0.06$ & $4.41 \pm 0.06$ & $0.19 $ \\
Claude 3.5 Sonnet    & $4.45 \pm 0.08$ & $4.25 \pm 0.08$ & $0.20 $ \\
Claude 4.5 Sonnet    & $4.58 \pm 0.06$ & $4.49 \pm 0.06$ & $0.09 $ \\
\hline
\end{tabular}
\caption{Cross-family LLM-judge re-evaluation on MS MARCO (Description). }
\label{tab:judge_families}
\end{table}

\section{Ablation Study}

\paragraph{Cost Analysis.}
We make two kinds of LLM calls in prompt optimization: inference calls to generate a prompt’s outputs and judge calls to compare outputs pairwise. Because our evaluation is empirical, we assess sample efficiency by comparing methods under the same judge-call budget.

Figure~\ref{fig:open_ended} shows that D-TS consistently finds better prompts than random sampling given the same number of judge calls. We also report mean wall-clock time on MS MARCO for SPO with its default 20 rounds and for PDO early-stopped at 10 rounds (Table~\ref{tab:wall_clock_time}). We use 10 rounds for PDO because Figure~\ref{fig:open_ended} indicates that by round 10, the prompt selected by PDO already surpasses SPO in test accuracy.

SPO is faster per run largely because it compares only two prompts per round and immediately eliminates the loser. However, without an explicit exploration–exploitation strategy, this procedure can be brittle under noisy comparisons, which helps explain its weaker performance.

Practically, as discussed in Section~\ref{preference_judge}, we control judge noise by using a strong but expensive model (e.g., Claude-4.5) to guide the search, and then deploying the optimized prompt on a lower-cost model (e.g., Llama-3.3) that serves production traffic. In this setting, D-TS is especially effective because it achieves better prompts with fewer expensive judge calls.
\begin{table}[htbp]
\centering
\scriptsize % shrink font just for table
\setlength{\tabcolsep}{1.75pt} % very narrow spacing
\renewcommand{\arraystretch}{1.3}
\begin{tabular}{lcccc}
\hline
\textbf{Method} & \textbf{Description} & \textbf{Entity} & \textbf{Numeric} & \textbf{Location} \\
\hline
PDO & \(\mathbf{4.39}\) & \(\mathbf{3.39}\) & \(\mathbf{3.83}\) & \(\mathbf{3.10}\) \\
SPO                           & 6.78 & 6.74 & 6.17 & 7.45 \\
\hline
\end{tabular}
\caption{Mean wall-clock time (minutes) on MS MARCO across four prompt tasks.}
\label{tab:wall_clock_time}
\end{table}

\begin{figure}[htbp]
  \centering
  \includegraphics[width=\linewidth]{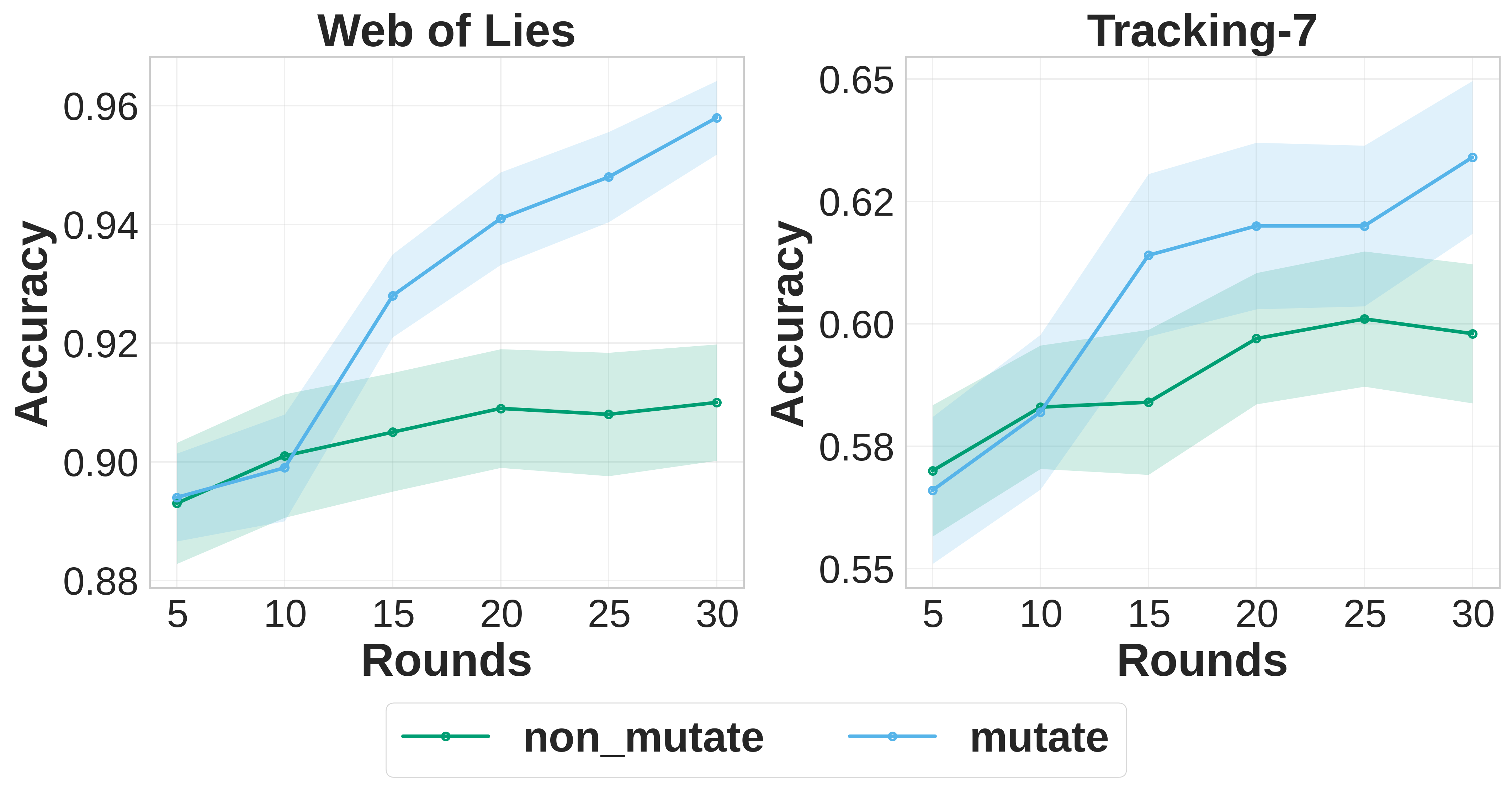}
  \caption{Effect of top-performer–guided mutation on \textit{Web of Lies} (left) and \textit{Tracking-7} (right).}
  \label{mutate}
\end{figure}

\paragraph{Prompt Mutation.}
PDO incrementally expands the prompt pool by mutating top-performing prompts. We visualize its effect on \textit{Web of Lies} and \textit{Tracking-7}. In the \emph{mutate} variant, at rounds 10 and 20 we prune the 10 lowest–Copeland-score prompts and replace them with 10 new candidates generated by mutating the current Copeland leader. The \emph{non\_mutate} baseline keeps the original 20 prompts fixed throughout. At each round, we report the ground-truth accuracy of the current Copeland leader. As shown in Figure~\ref{mutate}, both methods behave similarly before round 10, but mutation yields consistently higher accuracy afterward, surpassing the limits of the fixed pool and navigating toward better prompt regions.

We additionally report results across all 16 BBH tasks in Table \ref{tab:mutation_ttest}. In total, 8 out of 16 tasks show statistically significant gains under mutation, and mutation never underperforms relative to the non-mutation baseline.

\begin{figure}[htbp]
    \centering
    \includegraphics[width=\linewidth]{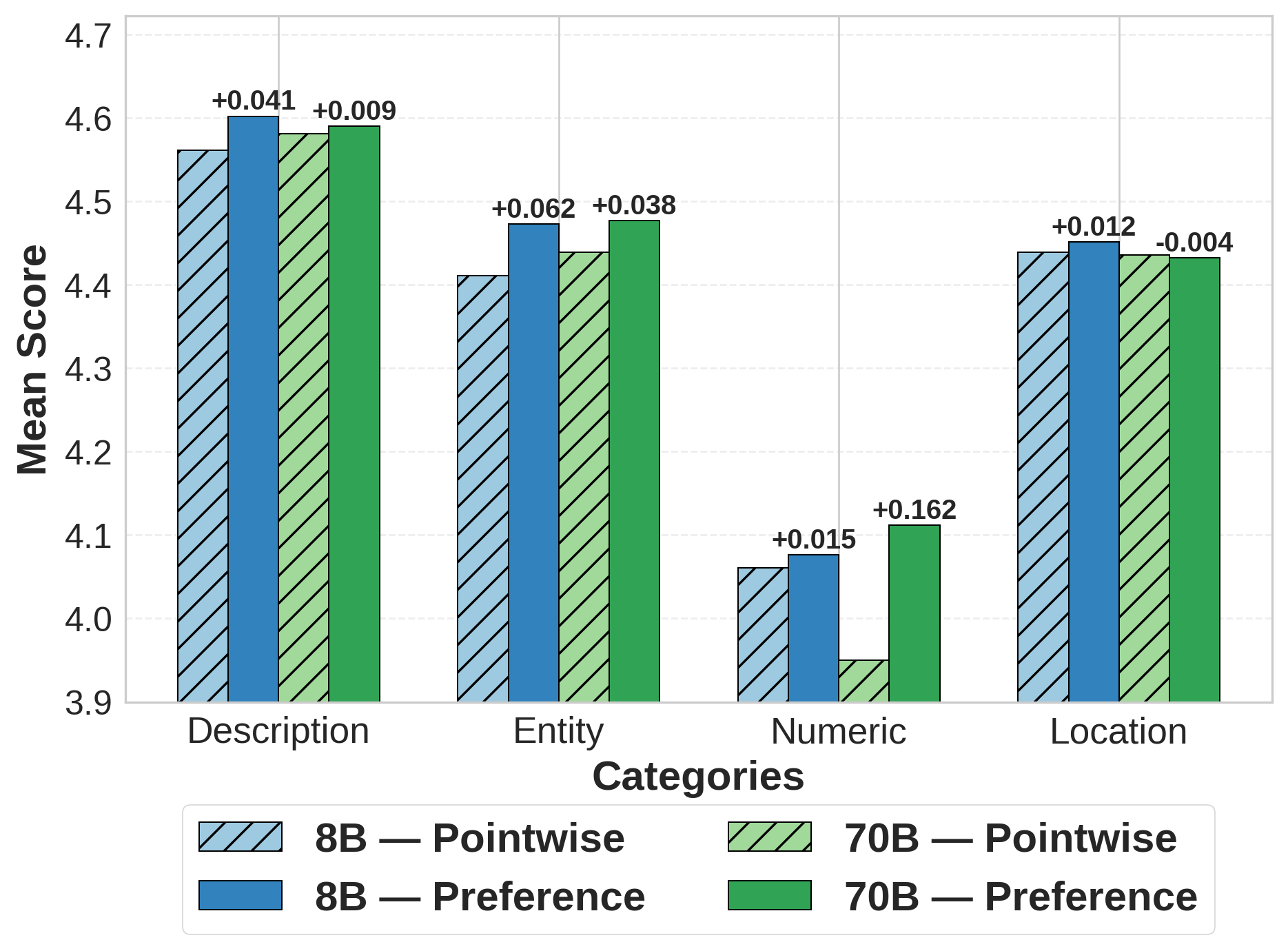}
    \caption{Comparison of pairwise preference used in PDO and pointwise scoring for prompt selection on MS-MARCO. Results indicate that pairwise preference leads to higher-performing prompts across model sizes (7 out of 8 cases).}
    \label{preference}
\end{figure}

\paragraph{Pairwise Preference vs. Pointwise Scoring.} PDO relies on the \emph{pairwise preferences} of an LLM judge to select the winning prompt. To highlight the benefit of this formulation, we compare it against a \emph{pointwise scoring} judge, where each prompt is evaluated independently on the development set by assigning a numeric score (1–5) without access to ground-truth answers. The prompt with the highest average score is then selected. We evaluate both strategies on the same fixed pool of 50 candidates from MS-MARCO in \ref{main_result} and report the test mean score of the selected prompt. Figure~\ref{preference} shows that preference judgement consistently outperforms pointwise scoring in 7 of 8 model–task combinations across two judge models. Overall, these results are consistent with the LLM-as-judge literature: pairwise comparisons yield more stable decisions than direct pointwise scoring by reducing dependence on calibration \cite{liu2024pairwiseCOLM}. 

\paragraph{Analysis of Exploration Strategies}\label{other_db}

\begin{table*}[htbp]
\centering
\scriptsize
\renewcommand{\arraystretch}{1.15}

\label{tab:exploration_comparison_all_rounds}
\begin{tabular}{c l c c c c}
\hline
\textbf{Round} & \textbf{Method} & \textbf{Description} & \textbf{Entity} & \textbf{Numeric} & \textbf{Location} \\
\hline

\multirow{6}{*}{5}
& D-TS           & \textbf{4.52} & \underline{4.38} & \textbf{4.00} & \textbf{4.32} \\
& Self-Sparring  & \underline{4.50} & \textbf{4.40} & \textbf{4.00} & \underline{4.30} \\
& RMED           & 4.47 & 4.37 & \underline{3.94} & 4.29 \\
& RUCB           & 4.46 & 4.32 & 3.92 & 4.26 \\
& MergeRUCB      & 4.44 & 4.34 & 3.88 & 4.28 \\
& Random         & 4.40 & 4.36 & 3.89 & 4.20 \\
\hline

\multirow{6}{*}{10}
& D-TS           & \textbf{4.57} & \underline{4.44} & \textbf{4.07} & \textbf{4.41} \\
& Self-Sparring  & \underline{4.56} & \textbf{4.46} & \underline{4.06} & \underline{4.40} \\
& RMED           & 4.53 & 4.42 & 4.02 & 4.39 \\
& RUCB           & 4.52 & 4.38 & 4.00 & 4.36 \\
& MergeRUCB      & 4.50 & 4.40 & 3.95 & 4.38 \\
& Random         & 4.45 & 4.41 & 3.97 & 4.29 \\
\hline

\multirow{6}{*}{15}
& D-TS           & \textbf{4.61} & \underline{4.48} & \textbf{4.11} & \textbf{4.45} \\
& Self-Sparring  & \underline{4.59} & \textbf{4.50} & \textbf{4.11} & \underline{4.43} \\
& RMED           & 4.57 & 4.45 & \underline{4.05} & 4.42 \\
& RUCB           & 4.55 & 4.40 & 4.03 & 4.39 \\
& MergeRUCB      & 4.53 & 4.42 & 3.98 & 4.41 \\
& Random         & 4.48 & 4.44 & 4.00 & 4.33 \\
\hline

\end{tabular}
\caption{Comparison of exploration strategies on MS-MARCO across optimization rounds in PDO. We report the mean test score of the current Copeland leader over rounds, selected by different exploration strategies. For each \{round, dataset\}, the best score is shown in \textbf{bold} and the second best is \underline{underlined}.}
\label{exploration_comparison_all_rounds}
\end{table*}

The dueling-bandit literature spans several algorithmic families. A useful high-level division is between \emph{non-stochastic} index policies, which select the next duel deterministically based on confidence or divergence indices, and \emph{stochastic} policies, which randomize comparisons via posterior sampling. To justify the use of D-TS as the exploration strategy in PDO, we conduct additional experiments comparing two UCB-style index methods (RUCB~\cite{RUCB2014} and MergeRUCB~\cite{zoghi2015mergerucb}), a KL-divergence index method (RMED~\cite{komiyama2015rmed}), and two posterior-sampling methods (D-TS~\cite{wu2016double} and Self-Sparring~\cite{sui2017selfsparring}).

Table~\ref{exploration_comparison_all_rounds} shows monotonic improvement from Round~5 to~15 for all methods. Under the same judge budget, posterior-sampling approaches are consistently strongest: D-TS achieves the best performance in most (round, task) cells, while Self-Sparring remains highly competitive and is the top-performing method on \textsc{Entity} at every reported round. RMED consistently falls between the TS-based methods and the UCB-style baselines—outperforming RUCB and MergeRUCB but remaining slightly below posterior sampling—while random selection performs worst overall.

We conjecture that this performance gap reflects characteristics of the prompt-optimization setting: a short optimization horizon with noisy preference feedback, where many prompt pairs are difficult to reliably separate and the effective notion of “best” aligns more closely with overall win frequency than with a clearly dominant Condorcet winner. In such regimes, posterior sampling can act as an \emph{anytime}, uncertainty-aware heuristic that continues to explore multiple plausibly strong prompts without over-investing in resolving near-ties. By contrast, index-based methods (both UCB- and divergence-based) may allocate a larger fraction of a limited budget to shrinking confidence sets or satisfying information constraints, which can be less effective when the horizon is small and the judge signal is noisy. 

\section{Related Work}
% \paragraph{Supervised APO Methods.} Traditional prompt optimization methods rely heavily on supervised signals from labeled validation sets, where the typical  pipeline involves generating, mutating, and
% scoring prompts \cite{zhou2022ape,fernando2023promptbreeder,pryzant2023protegi, guo2023evoprompt,schulhoff2024promptingsurvey}. Out 

% or carefully designed reward functions \cite{zhou2022ape,fernando2023promptbreeder,pryzant2023protegi, guo2023evoprompt, Ma2024AreLL, pryzant2023protegi, yang2024llm_optimizers, Kong2024PRewritePR}. InstructZero \cite{chen2024instructzero} and the methods proposed by Sabbatella et. al \cite{Sabbatella2024PromptOI, candelieri2023boing} employ Bayesian Optimization (BO) to search the combinatorial space of prompts, but rely on ground-truth labels to evaluate prompt performance. Similarly, PromptBreeder \cite{fernando2023promptbreeder} and EvoPrompt \cite{guo2023evoprompt} use evolutionary algorithms to generate mutated prompts from the fittest parents (prompts), but both calculate fitness scores from training datasets rather than operating in label-free settings. 

% Prior work on prompt optimization has largely relied on supervised signals, such as labeled validation sets or carefully designed reward functions \cite{zhou2022ape,fernando2023promptbreeder,pryzant2023protegi,guo2023evoprompt,Ma2024AreLL,yang2024llm_optimizers,Kong2024PRewritePR}.

\paragraph{APO Methods: from Supervised to Label-free.}
Traditional prompt optimization methods rely heavily on supervised signals from labeled validation sets, where the typical pipeline involves generating, mutating, and scoring prompts \cite{zhou2022ape,fernando2023promptbreeder,pryzant2023protegi,guo2023evoprompt,schulhoff2024promptingsurvey}. More recently, studies have reduced supervision requirements \cite{Xiang2025SPO,chen-etal-2024-prompt,Madaan2023SelfRefineIR,Cheng2023BlackBoxPO,zhan2022pace}. Notably, Self-Supervised Prompt Optimization (SPO) \cite{Xiang2025SPO} eliminates external references by using output-vs-output comparisons, iteratively generating, executing, and comparing prompts through pairwise LLM judgments. However, SPO follows a \emph{greedy hill-climbing loop} without a principled exploration–exploitation strategy, limiting its ability to allocate comparisons efficiently and to robustly identify high-performing prompts. 
% Test-time Preference Optimization \cite{li2025testtime} aligns LLM outputs during inference using textual gradient updates informed by reward signals from another LLM, but relies on raw numeric reward values rather than the more reliable pairwise comparisons from LLM judges \cite{zheng2023mtbench,liu-etal-2023-geval,gu2024llmjudgeSurvey}.

% \paragraph{Hybrid Methods with Minimal Human Feedback.} Several approaches have explored efficient incorporation of human guidance through preference feedback. Prompt Optimization with Human Feedback (POHF) \cite{lin2025prompt} adapts reinforcement learning with human feedback principles to prompt optimization, training neural networks on historical pairwise preferences to assign prompt scores. Lu et al. \cite{Lu2024PromptOW} propose a similar framework using pairwise comparisons with Elo ratings, allowing ties and providing robustness against single poor performances. While these methods reduce annotation burden compared to traditional supervised approaches, they still require human judges in the loop.

\paragraph{Bandit-Based Prompt Optimization.} The connection between prompt optimization and multi-armed bandits (MAB) has recently gained attention, as prompts can naturally be viewed as arms. OPTS \cite{ashizawa-etal-2025-bandit} formulates prompt strategy selection as a bandit problem with Thompson sampling. TRIPLE \cite{Shi2024EfficientPO} casts prompt optimization as fixed-budget best-arm identification, assuming a predefined set of prompts with known scores. However, both approaches require labeled validation sets for scoring. APOHF \cite{lin2025prompt} connects preference-based prompt optimization to dueling bandits, but assumes human-annotated pairwise preferences that are impractical at scale. PDO retains the dueling-bandit formulation while replacing human preferences with LLM-judge comparisons to align with realistic prompt optimization settings.

\section{Conclusion}

In this paper, we introduce PDO, a label-free prompt optimization method by framing the problem as a dueling bandit guided by LLM preference feedback. By combining Double Thompson Sampling (D-TS) with top-performer-guided mutation, PDO adaptively searches for informative prompt comparisons to identify high-performing prompts. Experiments on BBH and MS-MARCO datasets show that PDO consistently achieves competitive performance against baseline methods across tasks. We analyze judge noise and potential circularity, and demonstrate that PDO’s gains persist under alternative judge families. More broadly, our results suggest that LLM pairwise preference feedback can serve as a practical supervision signal for prompt optimization when labeled data are unavailable. Beyond the single-turn settings considered here, PDO provides a flexible foundation for extending label-free prompt optimization to more complex scenarios, such as multi-turn interactions and prompt sets for agentic systems.

\section*{Limitation}
While our dueling-bandit framework provides a practical and sample-efficient approach to prompt optimization without ground-truth labels, it also has inherent limitations. The capability of the preference judge plays a critical role: depending on task complexity, the judge may be noisy. Stronger foundation models tend to yield more reliable preferences but incur substantially higher cost, whereas smaller or less specialized models may introduce greater noise, particularly on domain-specific tasks. Because the algorithm optimizes for the judge’s notion of quality rather than the true task metric, it may favor stylistic patterns that align with the judge’s preferences.

To mitigate these concerns, we include a detailed discussion and ablation studies on LLM judge design, along with a small-scale human evaluation, in Section~\ref{judge} and Appendix~\ref{appendix_judge_design}. However, we were unable to conduct broader experiments across a wider range of tasks. Despite these challenges, our formulation provides a principled and scalable starting point for label-free prompt optimization using preference feedback, and we hope it motivates future work on improving the alignment between LLM judges and task objectives.

\bibliography{main}
\clearpage
\appendix

% \section{Appendix}
\label{sec:appendix}

\section{Small-scale Human Agreement Check}\label{human_agreement}

In addition to the cross-judge evaluation in Table \ref{tab:judge_families}, we examine whether the original Llama~3.3 MS-MARCO judge aligns with human preferences. We select 20 evaluation examples (5 from each of the four MS-MARCO tasks) for which the cached PDO and SPO outputs receive different Llama~3.3 scores under the 1–5 rubric (e.g., 5 vs.\ 4). We then recruit 10 non-author volunteer raters (graduate-student participants from a university population) to perform a blinded pairwise comparison: for each example, raters are shown the input and two anonymized outputs (PDO vs.\ SPO), with left/right order randomized and prompt identities hidden, and are asked to choose the output they prefer under the same criteria as the Llama judge (accuracy, completeness, relevance, and clarity). 

\begin{figure}[htbp]
  \centering
  \includegraphics[width=\linewidth]{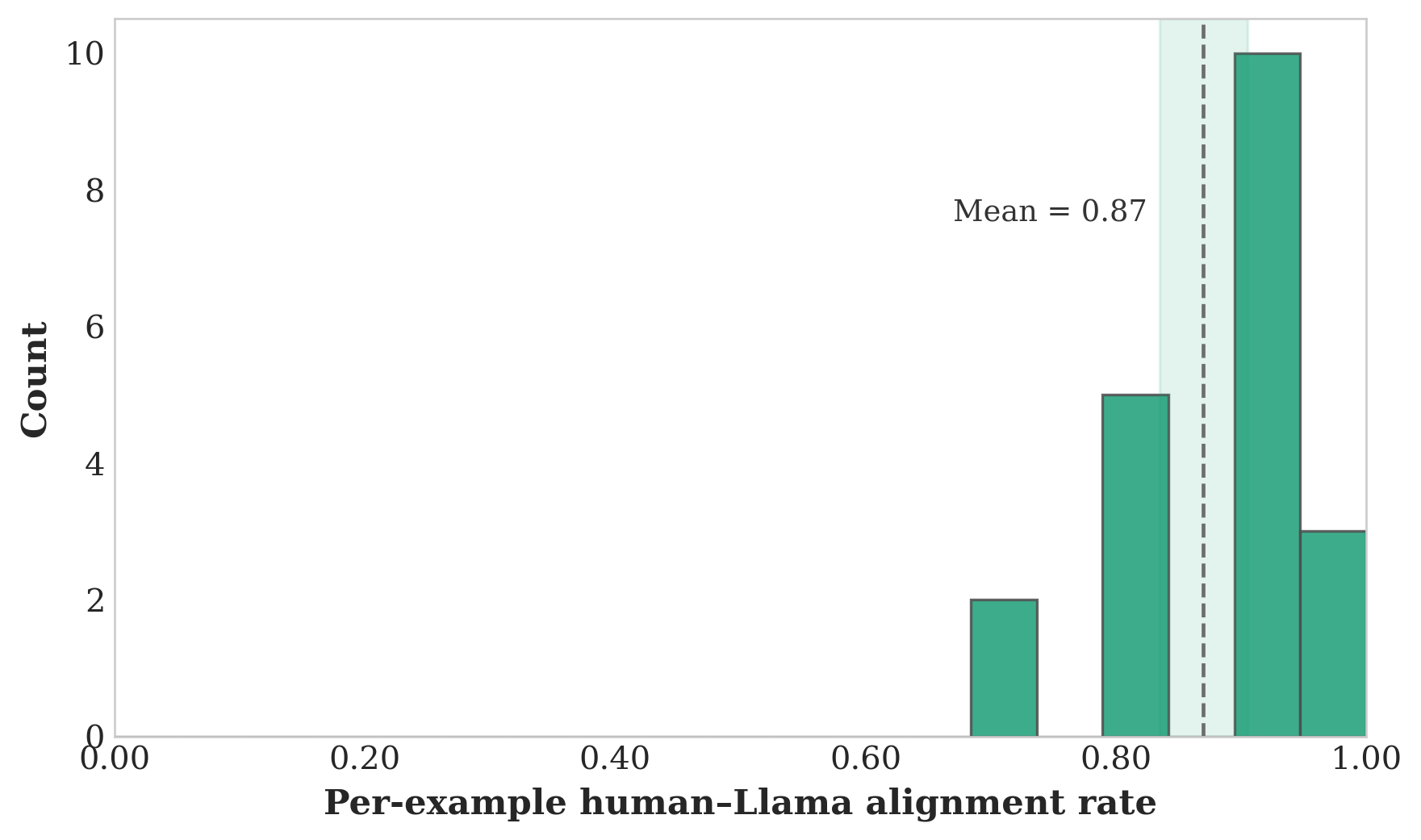}
  \caption{Histogram of per-example human–Llama alignment rates on 20 MS MARCO examples, where PDO and SPO outputs receive different Llama~3.3 evaluation scores.}
  \label{human_rater}
\end{figure}

For each example, we compute an alignment rate as the fraction of raters whose preferred output matches the direction implied by the Llama scores, and we summarize the distribution of alignment rates across the 20 examples in Figure \ref{human_rater}. Overall, this distribution suggests that the Llama-based evaluation signal is broadly consistent with human preferences on the subset of cases most relevant to distinguishing PDO from SPO, with a mean alignment ratio of 0.87.

\section{Experiment Details} \label{Experiment_details}
\subsection{Hyperparameters} 
For the main experiments in Section~\ref{main_result}, we initialize with 20 candidate prompts for BBH and 50 candidate prompts for MS-MARCO. Each experiment runs for 30 rounds, with 25 duels per round. At rounds 10 and 20, we apply prompt mutation by selecting the top-3 prompts ranked by Copeland scores and generating 10 new prompts. At the same time, we prune the 10 lowest-ranked prompts by Copeland scores. We always select the prompt with the highest Copeland score as the winner, using the average win rate as a tiebreaker when multiple prompts share the same score. The D-TS parameter is fixed at $\alpha=1.2$ throughout all experiments.

\subsection{Prompt generation strategies} 
To generate the initial candidate prompts, we follow the approach in MiPROV2 \cite{MiPROV2}, which first uses an LLM to construct a dataset summary from demonstration examples. The prompt generation process then incorporates this dataset summary, additional unlabelled demonstration examples, and randomly selected prompt tips from a predefined set to encourage diversity. In the mutation stage, we create new variants of the top-performing prompt by applying prompt tips to guide local edits, such as changing the tone, adjusting the wording, or appending synthetic few-shot examples. 

\subsection{Baselines}
\paragraph{Unsupervised (label-free) methods}
\textbf{SPO} follows an iterative optimize--execute--evaluate loop. In each round, the method first samples $n_{\text{examples}}$ fresh examples and evaluates the current prompt by running one answer-generation LLM call per example. A revised prompt is then proposed using a single LLM call conditioned on the observed cases. Both the current and revised prompts are evaluated on the same set of examples via an additional $n_{\text{examples}}$ answer-generation calls. The two sets of outputs are subsequently compared using $n_{\text{judge\_trials}}$ pairwise LLM judge calls, where each judge observes both outputs with randomized order to mitigate positional bias and votes for the preferred prompt. The revised prompt is accepted if it receives a majority of the judge votes; otherwise, the current prompt is retained. Overall, each round incurs $2n_{\text{examples}} + n_{\text{judge\_trials}} + 1$ LLM calls. 

With the default setting $n_{\text{examples}}=3$ and $n_{\text{judge\_trials}}=3$, this corresponds to 10 calls per round, and we run 20 rounds to obtain the main results reported in Section~4.

\paragraph{Supervised methods} For supervised APO methods, the typical pipeline involves generating, mutating, and scoring prompts. \textbf{APE} frames prompt design as black-box optimization, generating candidate instructions from input–output demonstrations and refining them through iterative search. \textbf{OPRO} treats the LLM itself as the optimizer: a meta-prompt with task descriptions and prior results guides the model to propose and evaluate new candidates in an iterative loop. \textbf{Breeder} applies an evolutionary approach, jointly evolving task-prompts and mutation-prompts through LLM-driven mutation and selection.
\subsection{Datasets}

\paragraph{BIG-bench Hard (BBH).}
BBH \citep{suzgun2022BBH} is a curated subset of BIG-bench consisting of 23 reasoning-intensive tasks. It is commonly used to stress multi-step reasoning and symbolic manipulation. In our experiments, we evaluate on 16 BBH multiple-choice tasks, where LLaMA-3.3-70B shows non-trivial sensitivity to instruction prompts.
\paragraph{MS MARCO.}
MS MARCO \citep{bajaj2016msmarco} is a large-scale question answering dataset built from real Bing search queries, paired with human-written answers and linked passages. It supports QA, passage ranking, and related IR/NLP tasks. In our setting, we focus on four task categories—Description, Entity, Numeric, and Location—and adopt a 1--5 integer scoring scheme from an LLM judge that compares model outputs against the dataset’s ground-truth answers.

\section{Discussion on Evaluation Scope}
We evaluate PDO on BBH and MS MARCO, which capture two practically important prompt-optimization regimes. BBH covers a broad set of real-world, prompt-based instruction-following \emph{classification} tasks in a cold-start setting, where labeled data are scarce and costly to obtain at the outset. This setting is increasingly relevant in industry workflows such as content moderation \cite{spotify2025policyasprompt}, where many systems are transitioning from fully supervised ML/DL pipelines to prompt-based solutions whose performance is largely determined by the underlying policy prompt. 

% \cite{spotify2025policyasprompt}.

In contrast, MS MARCO represents a single-turn instruction/QA setting, mirroring scenarios where practitioners optimize a system prompt for a locally deployed LLM application. In such cases, tasks are often open-ended and evaluation criteria can be subjective, making an LLM judge a natural and scalable choice for preference-based comparisons.

We admit that extending PDO to multi-turn interactions, as well as jointly optimizing \emph{sets} of prompts (e.g., for multi-agent systems), is a valuable and important direction for future work.

\clearpage 

\begin{table*}[htbp]
\centering
\scriptsize
\setlength{\tabcolsep}{3pt}%
\renewcommand{\arraystretch}{1.2}
\begin{tabular}{l|cccccccc}
\hline
\textbf{Method} & \textbf{Causal} & \textbf{Date} & \textbf{DisambigQA} & \textbf{Formal} & \textbf{Geometric} & \textbf{Hyperbaton} & \textbf{Logical-5} & \textbf{Logical-7} \\
\hline
APE            & \(0.680 \std{0.044}\) & \(0.892 \std{0.019}\) & \(0.730 \std{0.043}\) & \(\underline{0.747} \std{0.022}\) & \(0.670 \std{0.072}\) & \(\underline{0.940} \std{0.013}\) & \(\mathbf{0.822} \std{0.041}\) & \(0.721 \std{0.035}\) \\
OPRO          & \(\underline{0.682} \std{0.044}\) & \(\underline{0.910} \std{0.022}\) & \(0.734 \std{0.038}\) & \(0.728 \std{0.026}\) & \(0.569 \std{0.048}\) & \(0.932 \std{0.021}\) & \(0.774 \std{0.030}\) & \(0.718 \std{0.031}\) \\
Breeder  & \(\mathbf{0.683} \std{0.027}\) & \(0.898 \std{0.026}\) & \(\mathbf{0.745} \std{0.038}\) & \(0.746 \std{0.027}\) & \(\underline{0.684} \std{0.048}\) & \(0.932 \std{0.021}\) & \(0.778 \std{0.022}\) & \(\underline{0.724} \std{0.018}\) \\
PDO (ours)         & \(0.680 \std{0.033}\) & \(\mathbf{0.915} \std{0.024}\) & \(\underline{0.740} \std{0.040}\) & \(\mathbf{0.754} \std{0.022}\) & \(\mathbf{0.712} \std{0.058}\) & \(\mathbf{0.948} \std{0.017}\) & \(\underline{0.809} \std{0.028}\) & \(\mathbf{0.733} \std{0.024}\) \\
\hline
\textbf{Method} & \textbf{Navigate} & \textbf{Penguins} & \textbf{Salient} & \textbf{Snarks} & \textbf{Tracking-5} & \textbf{Tracking-7} & \textbf{Tracking-3} & \textbf{Web of Lies} \\
\hline
APE            & \(\underline{0.899} \std{0.024}\) & \(0.925 \std{0.025}\) & \(0.686 \std{0.033}\) & \(0.875 \std{0.033}\) & \(0.803 \std{0.041}\) & \(\underline{0.600} \std{0.050}\) & \(0.911 \std{0.041}\) & \(0.948 \std{0.020}\) \\
OPRO           & \(0.882 \std{0.032}\) & \(0.919 \std{0.025}\) & \(0.687 \std{0.023}\) & \(\underline{0.888} \std{0.021}\) & \(0.833 \std{0.083}\) & \(\mathbf{0.662} \std{0.073}\) & \(0.924 \std{0.035}\) & \(0.938 \std{0.045}\) \\
Breeder  & \(\mathbf{0.908} \std{0.016}\) & \(\underline{0.927} \std{0.018}\) & \(\mathbf{0.698} \std{0.024}\) & \(\mathbf{0.903} \std{0.019}\) & \(\mathbf{0.859} \std{0.049}\) & \(\underline{0.600} \std{0.042}\) & \(\underline{0.927} \std{0.028}\) & \(\underline{0.963} \std{0.019}\) \\
PDO (ours)         & \(0.890 \std{0.010}\) & \(\mathbf{0.933} \std{0.027}\) & \(\underline{0.695} \std{0.022}\) & \(0.874 \std{0.036}\) & \(\underline{0.847} \std{0.062}\) & \(\mathbf{0.662} \std{0.084}\) & \(\mathbf{0.946} \std{0.015}\) & \(\mathbf{0.979} \std{0.024}\) \\
\hline
\end{tabular}
\caption{We report test results for PDO when selecting the prompt with the highest development-set accuracy from the same candidate pools as in Table~\ref{Without_label}. This setting serves as a proxy for replacing the LLM preference judge with an oracle. We compare these results against supervised APO methods and find that PDO remains competitive with state-of-the-art prompt optimization baselines, achieving the best performance on 9 out of 16 tasks.
}
\label{With_label}
\end{table*}

\begin{figure*}[t]
\centering
\begin{subfigure}[b]{0.32\textwidth}
    \centering
    \includegraphics[width=\textwidth]{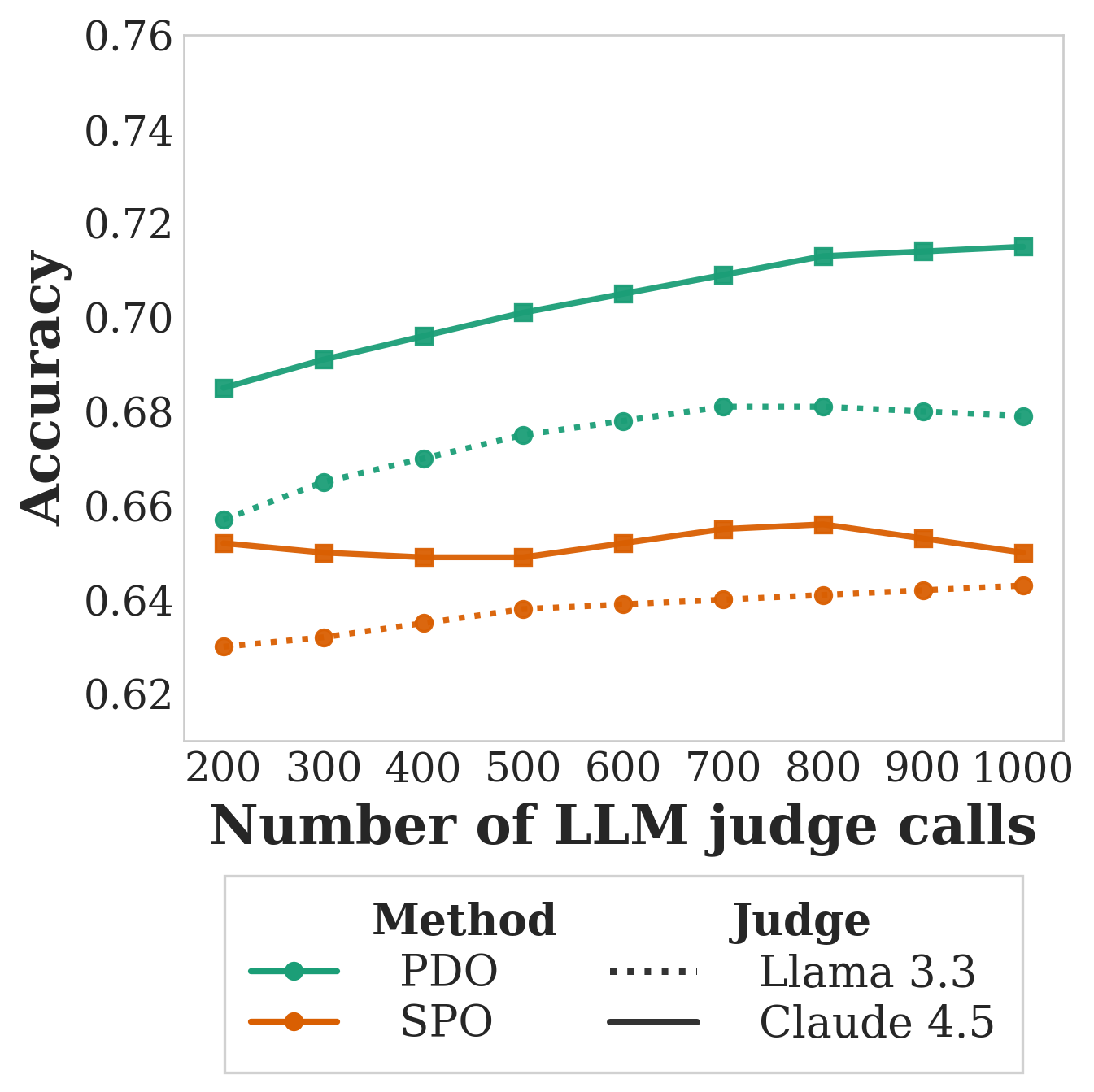}
    \caption{Causal}
    \label{fig:description}
\end{subfigure}\hfill
\begin{subfigure}[b]{0.32\textwidth}
    \centering
    \includegraphics[width=\textwidth]{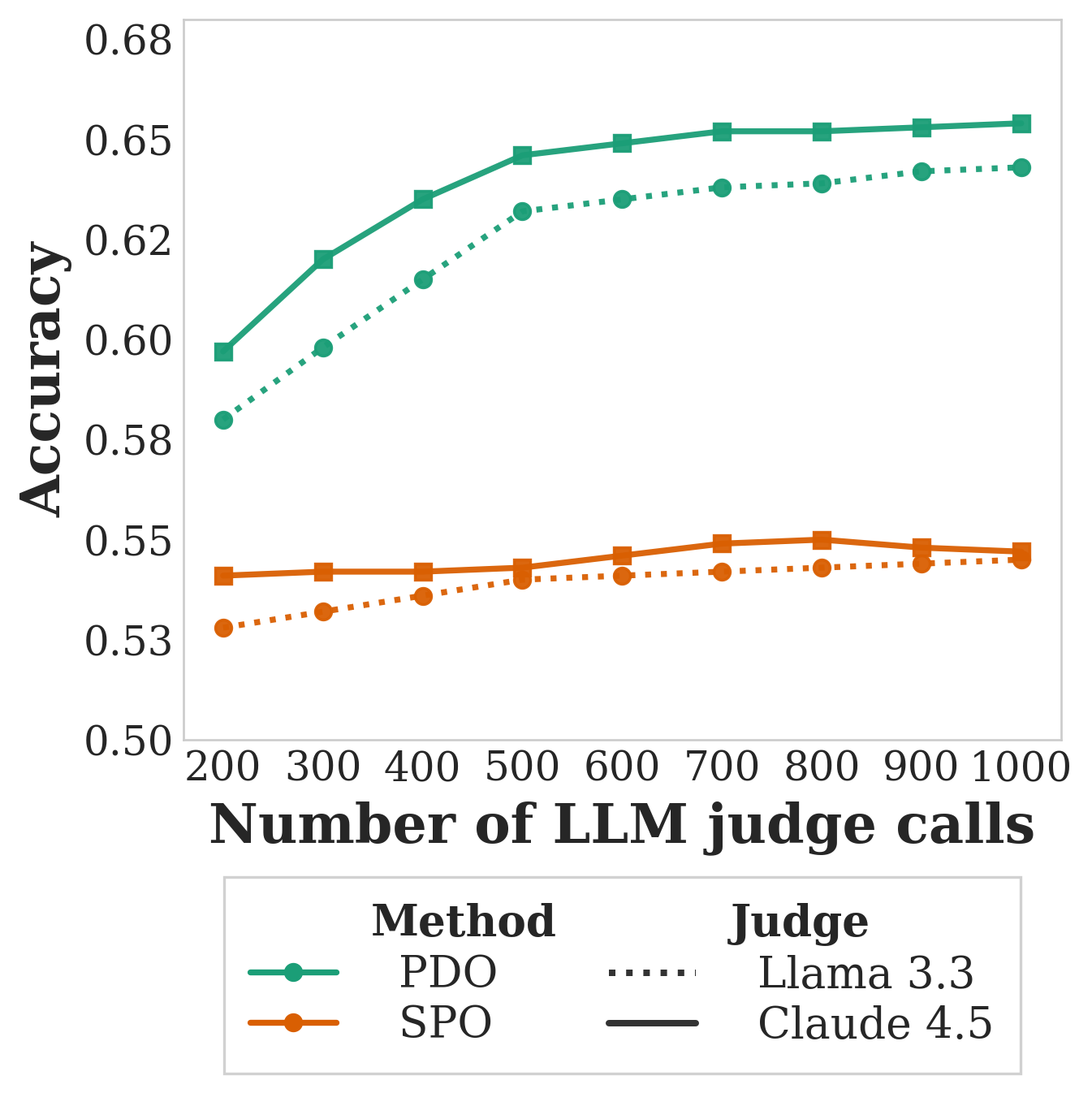}
    \caption{Tracking-7}
    \label{fig:entity}
\end{subfigure}\hfill
\begin{subfigure}[b]{0.32\textwidth}
    \centering
    \includegraphics[width=\textwidth]{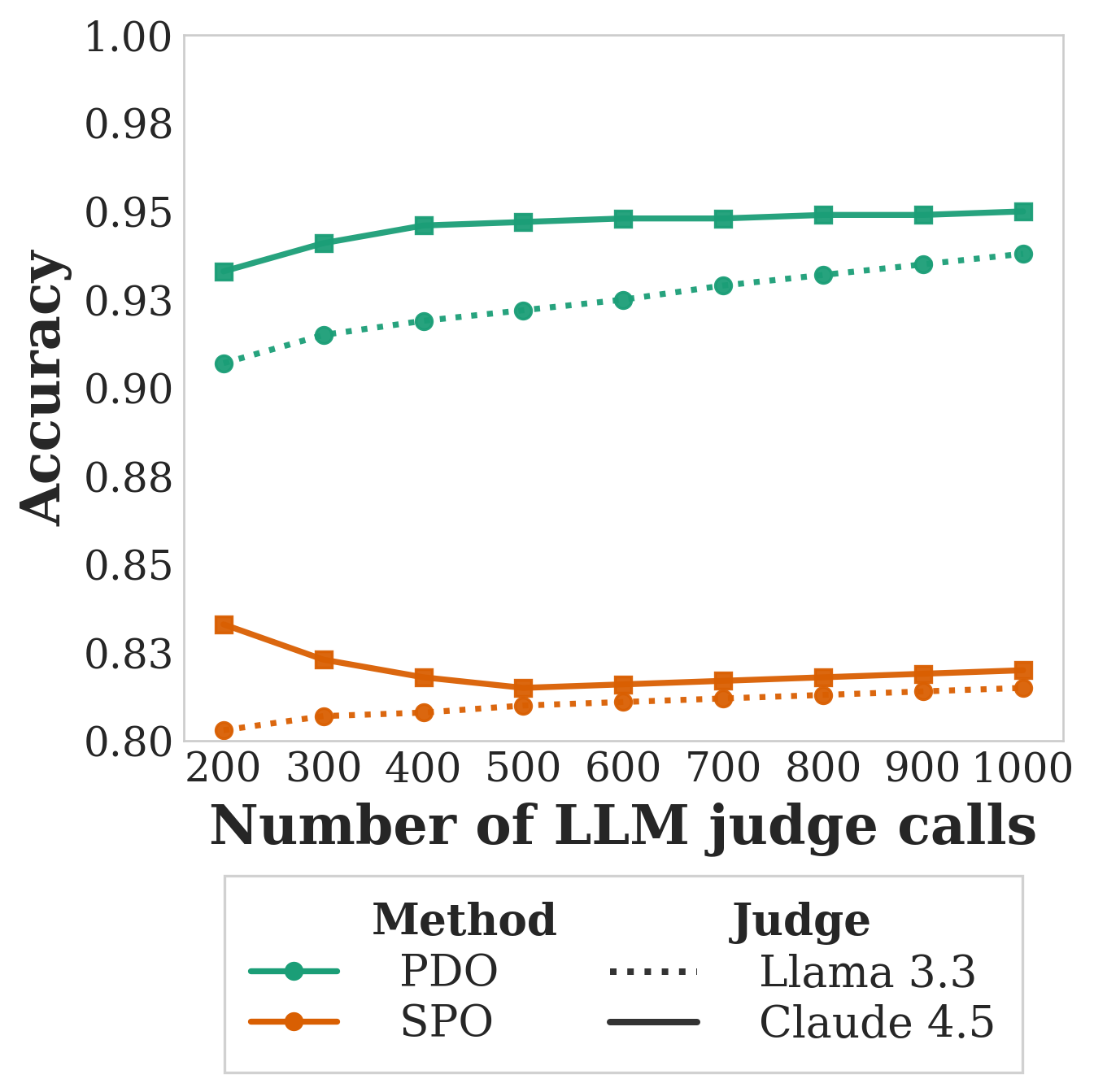}
    \caption{Web of Lies}
    \label{fig:numeric}
\end{subfigure}
\caption{Test performance when using Claude 4.5 vs. Llama 3.3 as the LLM preference judge on BBH tasks under the same judge-call budget. Using Claude 4.5 more reliably identifies higher-performing prompts; the improvement is largest on tasks where the Llama 3.3 judge is noisier (e.g., Causal in Figure 7a and Geometric in Figure \ref{claude_impact}).}
\label{fig:claude4.5more}
\end{figure*}
\clearpage

\begin{table*}[t]
\centering
\small
\setlength{\tabcolsep}{8pt}
\begin{tabular}{lcccc}
\hline
\textbf{Dataset} & \textbf{Non-mutate} & \textbf{Mutate} & \textbf{Diff.} & \textbf{$p$-value} \\
\hline
Causal        & 0.709 & 0.703 & -0.006 & $>0.05$ \\
Date          & 0.927 & 0.938 &  \phantom{-}0.011 & 0.0291 \\
DisambigQA    & 0.766 & 0.760 & -0.006 & $>0.05$ \\
Formal        & 0.759 & 0.767 &  \phantom{-}0.008 & $>0.05$ \\
Geometric     & 0.611 & 0.672 &  \phantom{-}0.061 & $9.7\times 10^{-5}$ \\
Hyperbaton    & 0.935 & 0.943 &  \phantom{-}0.008 & $>0.05$ \\
Logical-5     & 0.821 & 0.831 &  \phantom{-}0.010 & $>0.05$ \\
Logical-7     & 0.755 & 0.750 & -0.005 & $>0.05$ \\
Navigate      & 0.903 & 0.917 &  \phantom{-}0.014 & 0.0146 \\
Penguins      & 0.943 & 0.955 &  \phantom{-}0.012 & 0.0236 \\
Salient       & 0.719 & 0.711 & -0.008 & $>0.05$ \\
Snarks        & 0.863 & 0.876 &  \phantom{-}0.013 & $>0.05$ \\
Tracking-5    & 0.793 & 0.833 &  \phantom{-}0.040 & 0.0252 \\
Tracking-7    & 0.585 & 0.647 &  \phantom{-}0.062 & 0.0040 \\
Tracking-3    & 0.949 & 0.957 &  \phantom{-}0.008 & 0.0491 \\
Web of Lies   & 0.919 & 0.979 &  \phantom{-}0.060 & $1.5\times 10^{-6}$ \\
\hline
\end{tabular}
\caption{\textbf{Effect of prompt mutation on BBH.} We compare the final Copeland-winner prompt selected by PDO when mutation is disabled (\textit{Non-mutate}) versus when mutation is enabled (\textit{Mutate}). For each BBH task, we report the mean test accuracy over 10 independent runs. We perform a two-sample $t$-test per task to assess whether mutation yields a statistically significant accuracy change. Overall, 8 out of 16 tasks show significant improvement at $p<0.05$. We visualize representative tasks (Web of Lies and Tracking-7) in Figure \ref{mutate}. Tasks with small negative differences show only minor changes and are likely attributable to run-to-run noise; across all tasks considered, mutation does not materially degrade performance.}
\label{tab:mutation_ttest}
\end{table*}

\clearpage
\section{Further Discussion on LLM Judge Design for Closed-Ended Tasks}\label{appendix_judge_design}
The design of an LLM preference judge can vary depending on the type of task PDO tackles. In this section, we present the rationale and provide a detailed analysis of our judge design for multiple-choice BBH tasks. 

Since outputs in BBH are tied to fixed multiple-choice answers, the judging process naturally splits into two cases. (i) When the two prompts produce different answers, the judge’s role is straightforward: a capable judge should be able to identify the correct answer and select the corresponding prompt. (ii) When the two prompts produce the same answer, the situation is less clear. In this case, the answer alone cannot distinguish between the prompts, so we instead ask the judge to compare their reasoning chains. This design is well suited to BBH tasks, which are explicitly constructed to stress reasoning \citep{suzgun2022BBH}: the datasets involve multi-step deduction, object tracking, logical rules, or numerical reasoning, where strong performance correlates with the quality of intermediate steps. Our hypothesis is that examining the reasoning trace can reveal whether the model reached the correct answer through a coherent process or by shortcuts or luck. 

This formalizes our dual-judge approach, as discussed in Section~5.1, where the judge evaluates either the answer or the reasoning depending on the case. To enable this, we prompt the model to always produce both a reasoning chain and a final answer using JSON-guided decoding.

\subsection{Validating the Dual-Judge Approach}\label{appendix_validating_dual_judge}
To further probe the dual-judge approach, we conduct the following experiments. For each dataset, we select two prompts with different accuracy levels: prompt $A$ as the \textbf{higher-accuracy} prompt and prompt $B$ as the \textbf{lower-accuracy} prompt. We then test the judge in two scenarios: (i) whether prompt A is preferred when A is correct and B is wrong, and (ii) whether prompt A is preferred when both A and B produce correct answers. The results are reported in Figure~\ref{reason_answer}.

\paragraph{Result Analysis.}
For most BBH tasks, we find strong evidence that the LLM judge reliably selects the higher-accuracy prompt when $A$ is correct and $B$ is wrong. We also observe moderate evidence that when both prompts produce the same answer, the judge prefers $A$ based on the quality of its reasoning. Datasets where this preference is consistent under both \textbf{Answer} and \textbf{Reasoning} conditions (\textit{Hyperbaton}, \textit{Tracking-5}, \textit{Tracking-7}, and \textit{Web of Lies}) show superior performance in Table~\ref{Without_label}. On the other hand, when the judge is inconsistent in identifying the higher-accuracy prompt $A$ in either condition, performance degrades (\textit{Geometric}, \textit{Salient}).

\subsection{Weighted Preference Matrix Update}\label{appendix_discounted_preference}
\begin{figure*}
      \centering
  \includegraphics[width=0.87\linewidth]{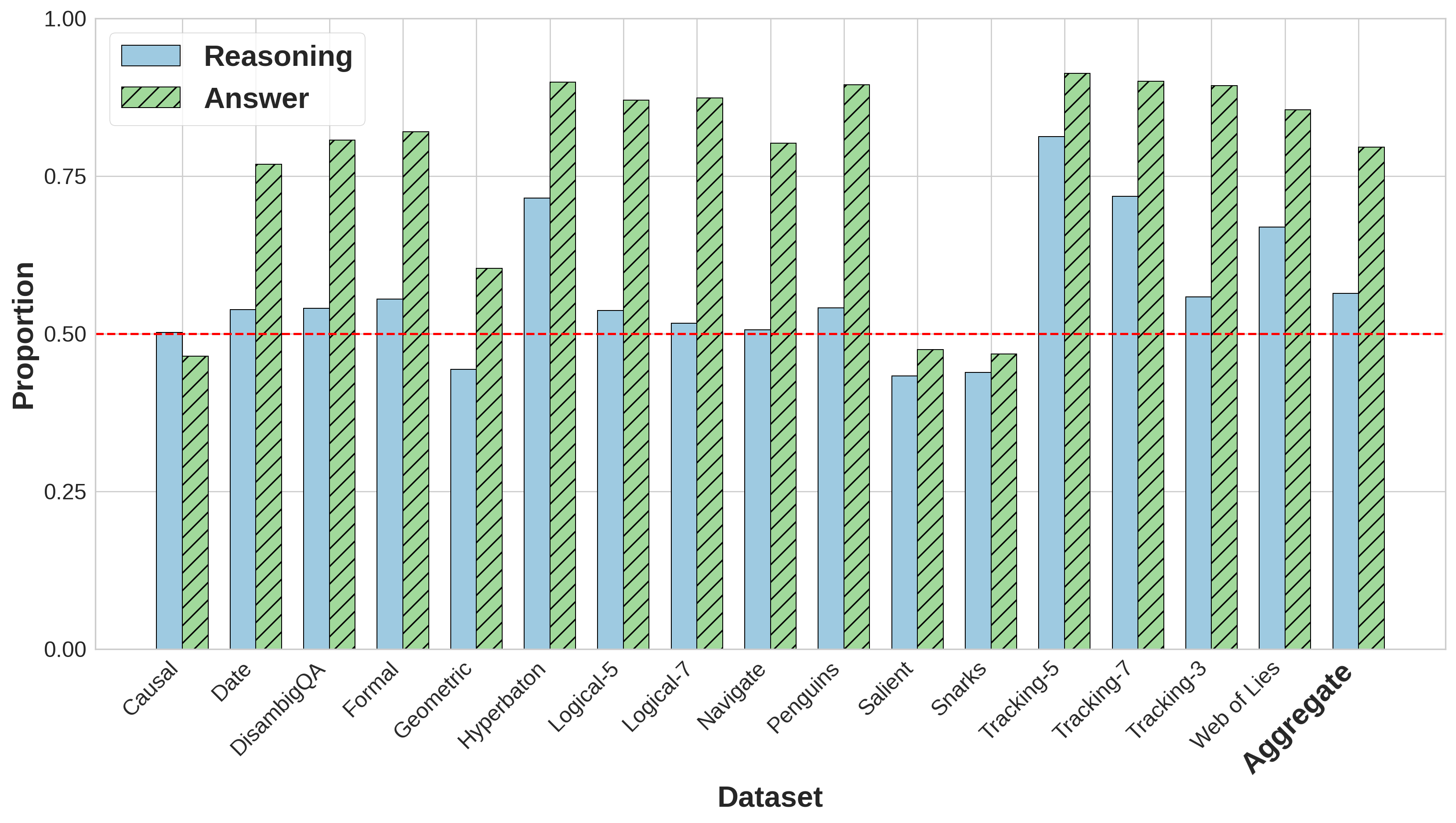}
  \caption{For each BBH dataset, we construct a prompt pair $(A,B)$, where $A$ is the higher-accuracy prompt and $B$ is the lower-accuracy prompt. We report the proportion of comparisons in which the judge prefers $A$. \textbf{Answer} (green) covers cases where $A$ is correct and $B$ is wrong; the judge evaluates by checking answers. \textbf{Reasoning} (blue) covers cases where both prompts produce the correct answer; the judge compares their reasoning chains. \emph{Aggregate} pools all examples across datasets.}
  \label{reason_answer}
\end{figure*}

\begin{figure*}
\centering
% --- First row ---
\begin{subfigure}{0.32\textwidth}
  \centering
  \includegraphics[width=\linewidth]{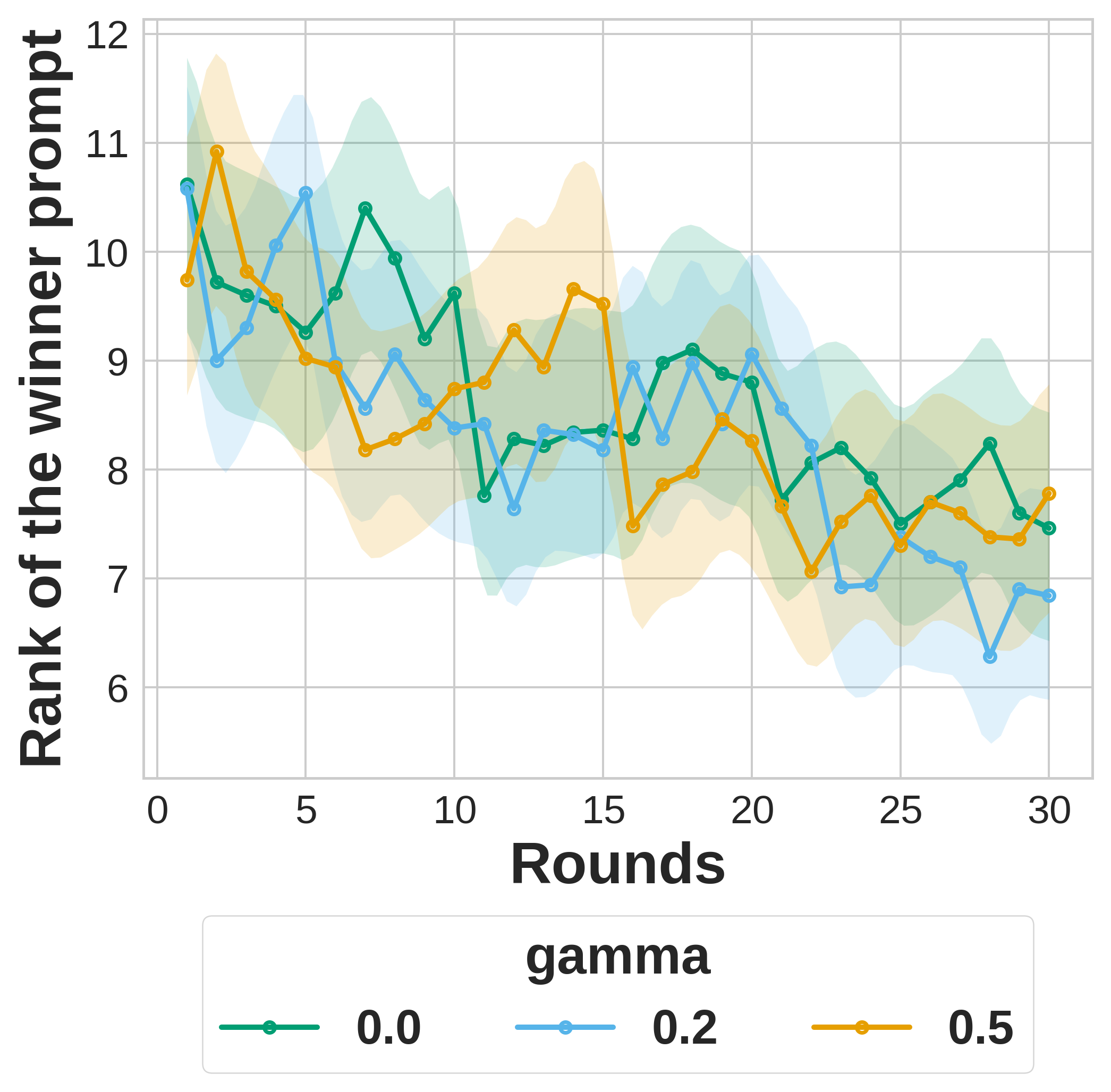}
  \caption{Logical-5}
  \label{fig:a}
\end{subfigure}
\hfill
\begin{subfigure}{0.32\textwidth}
  \centering
  \includegraphics[width=\linewidth]{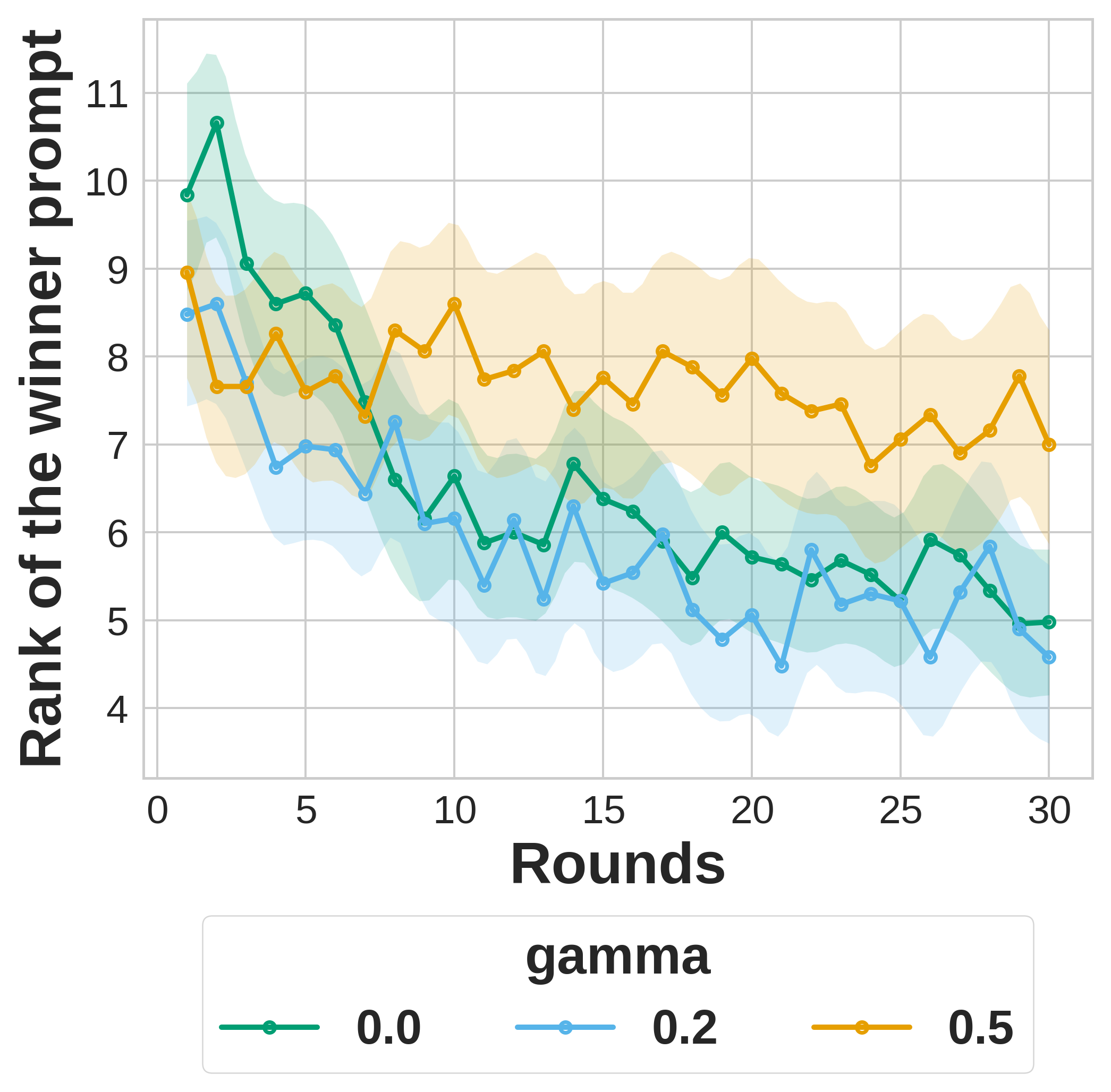}
  \caption{Hyperbation}
  \label{fig:b}
\end{subfigure}
\hfill
\begin{subfigure}{0.32\textwidth}
  \centering
  \includegraphics[width=\linewidth]{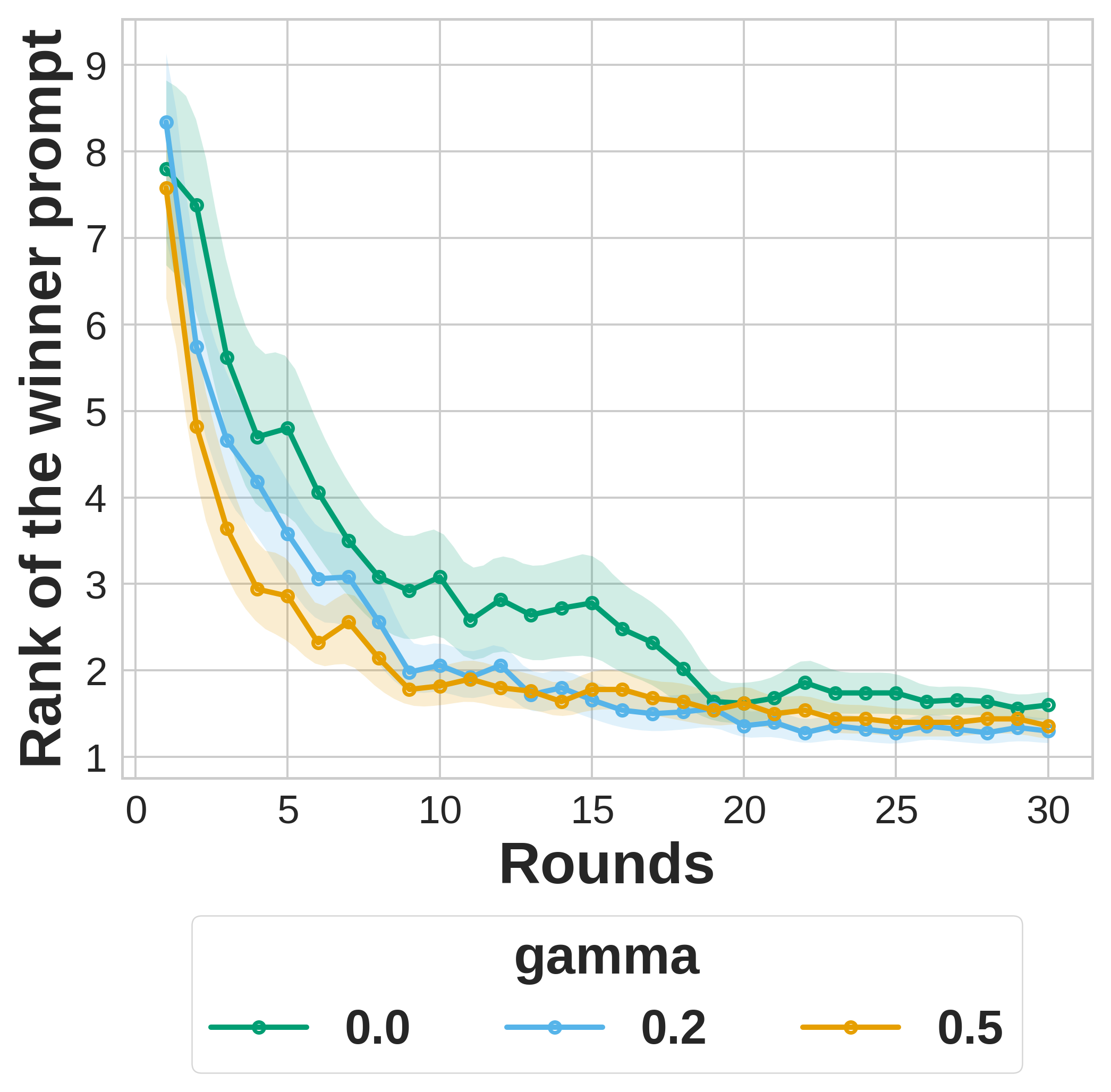}
  \caption{Tracking-7}
  \label{fig:c}
\end{subfigure}

\vspace{1em} % space between rows

% --- Second row ---
\begin{subfigure}{0.32\textwidth}
  \centering
  \includegraphics[width=\linewidth]{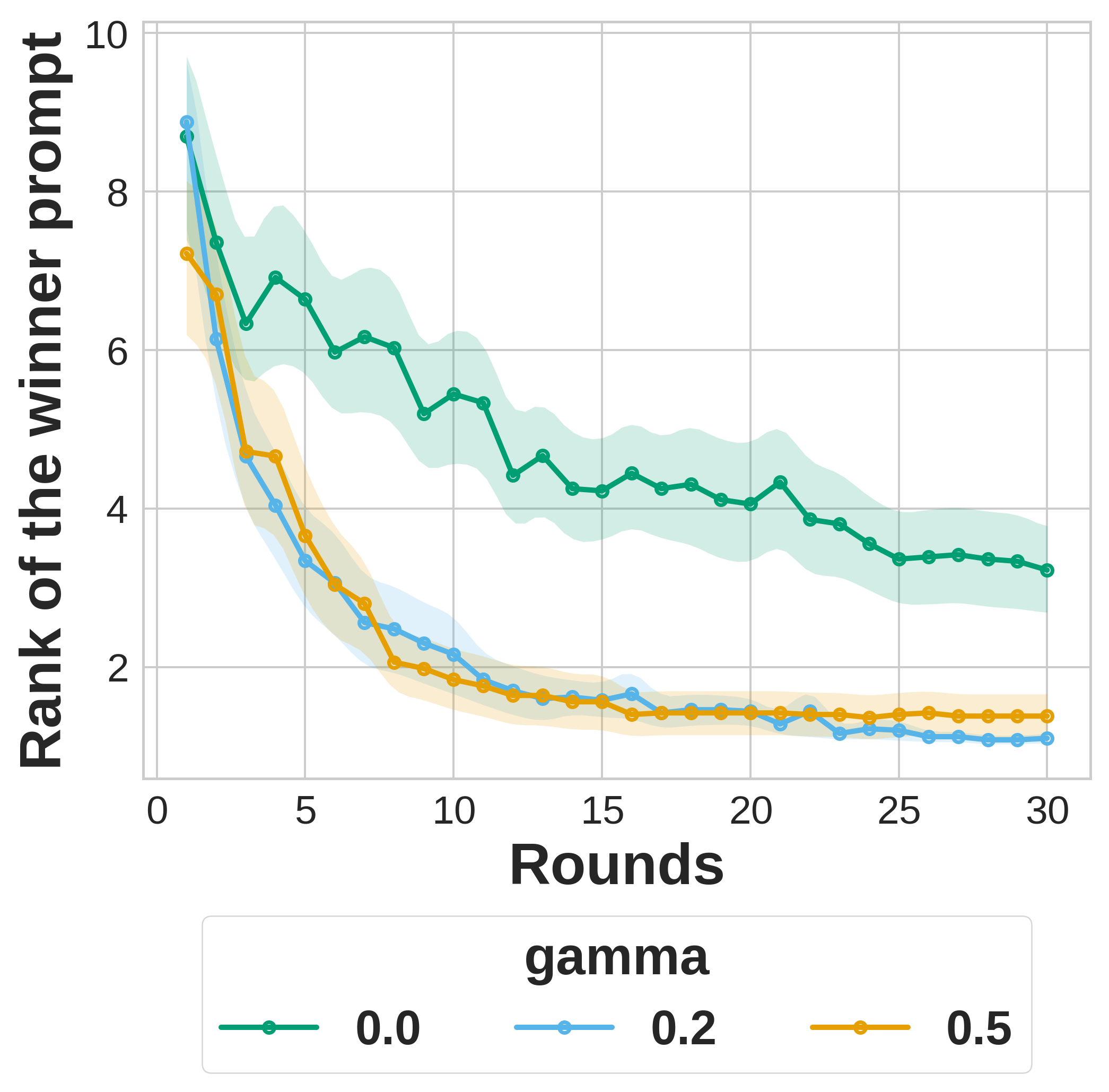}
  \caption{Web of lies}
  \label{fig:d}
\end{subfigure}
\hfill
\begin{subfigure}{0.32\textwidth}
  \centering
  \includegraphics[width=\linewidth]{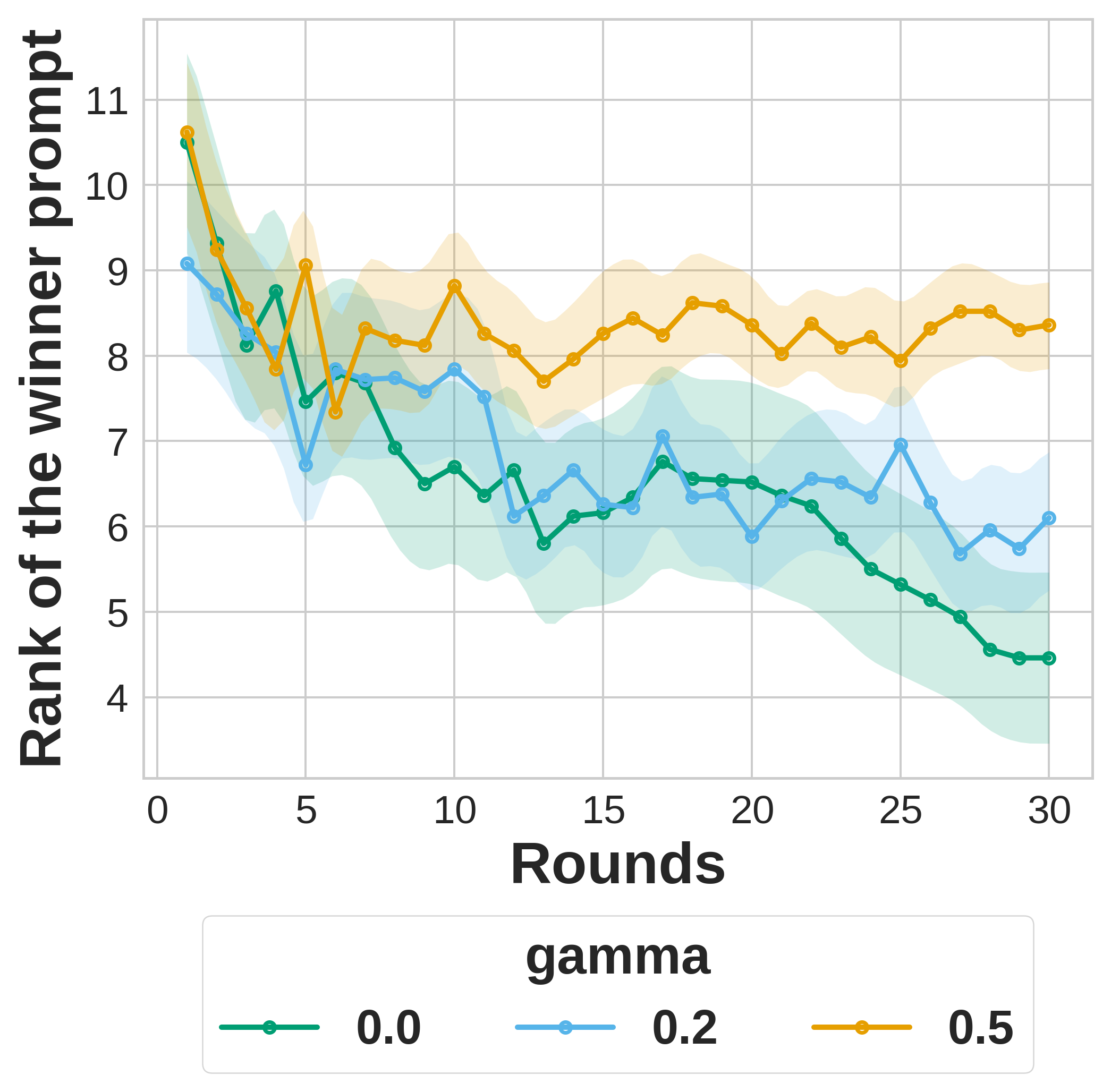}
  \caption{Formal}
  \label{fig:e}
\end{subfigure}
\hfill
\begin{subfigure}{0.32\textwidth}
  \centering
  \includegraphics[width=\linewidth]{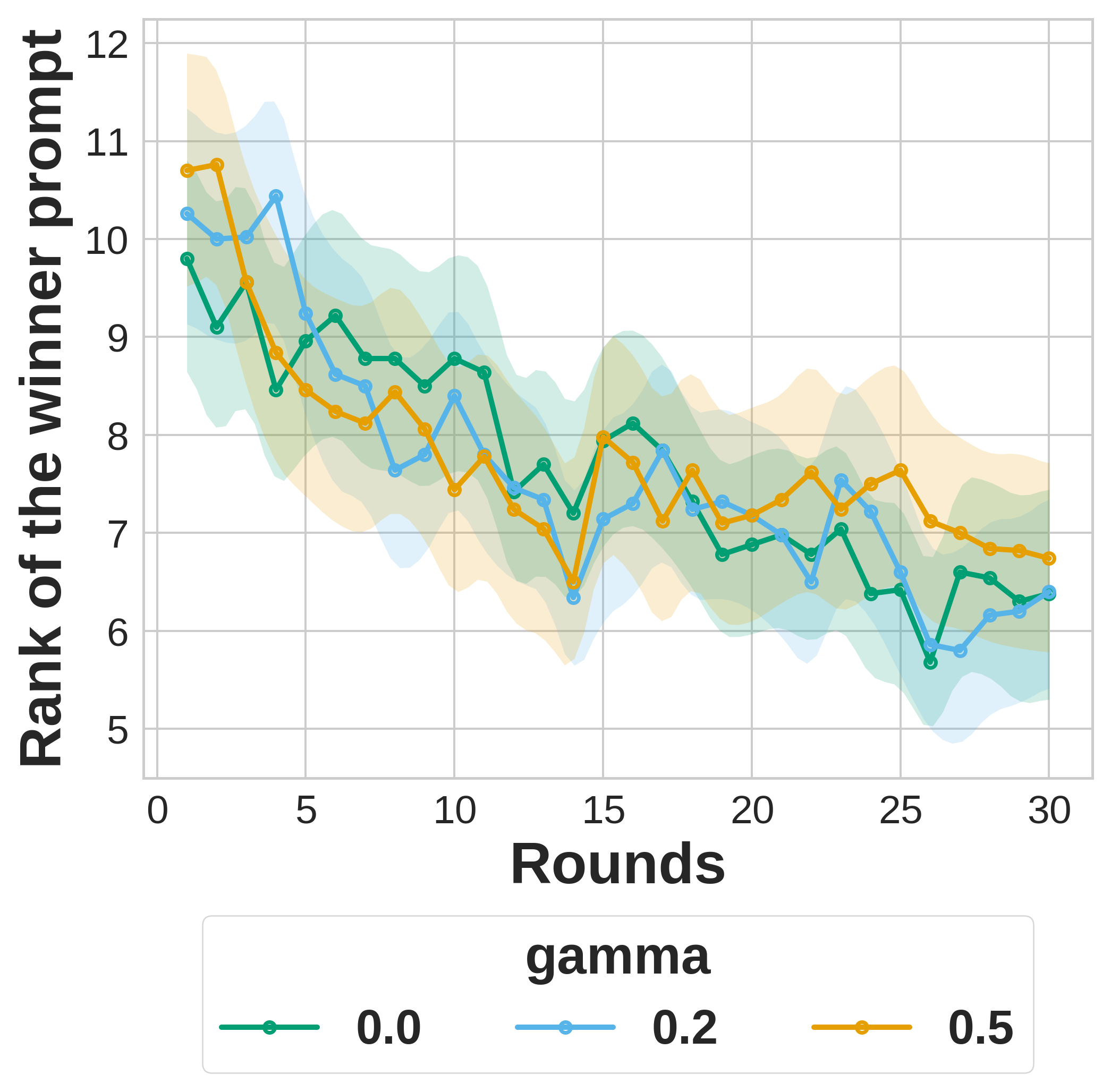}
  \caption{Date}
  \label{fig:f}
\end{subfigure}

\caption{Effect of the reasoning-discount $\gamma$ in the D-TS update across six BBH datasets. 
Each plot shows the ground-truth accuracy rank (lower is better) of the current Copeland leader over rounds. We fix $\gamma_{\text{Answer}}=0.5$ and ablate $\gamma_{\text{Reasoning}} \in \{0.0,\,0.2,\,0.5\}$. 
Results indicate that introducing a mild discount at $\gamma_{\text{Reasoning}}=0.2$ generally accelerates convergence and produces better final ranks overall compared to the undiscounted case $\gamma_{\text{Reasoning}}=0.5$.}
\label{gamma}
\end{figure*}

Motivated by the above analysis in Figure~\ref{reason_answer}, we consider a modification of the PDO algorithm for BBH tasks in which reasoning-based decisions, which are noisier than answer-based ones, are down-weighted so that they contribute less to the preference counts. Let $W_{ij}$ and $W_{ji}$ be the win counts used to form the Beta posteriors $\mathrm{Beta}(W_{ij}{+}1,\,W_{ji}{+}1)$ in D-TS.
For a duel decided by source $y \in \{\text{Answer}, \text{Reasoning}\}$, we introduce a margin parameter $\gamma_y \in [0,0.5]$ and update
\[
  (W_{ij}, W_{ji}) \leftarrow (W_{ij}+0.5{+}\gamma_y,\; W_{ji}+0.5-\gamma_y)
\]
whenever $i$ is preferred, with the symmetric rule when $j$ is preferred. This preserves one effective comparison per duel (sum increment $=1$) while shrinking the winner–loser gap to $2\gamma_y$ under noisier judgments.
We fix $\gamma_{\text{Answer}}=0.5$ and ablate $\gamma_{\text{Reasoning}} \in \{0.0,0.2,0.5\}$. 
Figure~\ref{gamma} reports, for each dataset, the ground-truth accuracy \emph{rank} of the Copeland leader over rounds. The results suggest that smaller values of $\gamma_{\text{Reasoning}}$ (e.g., $0.2$) often yield more stable progress than the full update $\gamma_{\text{Reasoning}}=0.5$ on tasks with noisier reasoning judgments, while tasks with consistent reasoning judgments remain robust across settings.

\paragraph{Takeaway.}
For BBH tasks, we use discounted updates as a simple mechanism to handle differences in judge reliability between answer-based and reasoning-based decisions in preference-based prompt optimization. More generally, for other tasks, LLM judges could be extended with adaptive adjustments that reflect reliability or with additional mechanisms to better capture uncertainty in judgments. Exploring such extensions is an interesting direction for future work.

\subsection{Case Study}
We conduct a case study of the judge's decisions in Table~\ref{tab:appendixA_case_study} on the BBH \emph{Web of Lies} task by comparing a lower-accuracy Prompt~A (0.852) with a higher-accuracy Prompt~B (1.000).
Two representative scenarios explain why the judge tends to prefer~B:
(i) when both prompts produce the correct answer, the reasoning-based judge compares the reasoning chains and favors~B's more concise and internally consistent rationale;
(ii) when the prompts disagree and~B's answer is correct, the answer-based judge correctly selects~B.

\begin{table*}[t]
\centering
\caption{Case study of Reasoning-based and Answer-based judging on Prompt~A (lower accuracy) and Prompt~B (higher accuracy) for selected questions from the \emph{Web of Lies} dataset in BBH.}
\label{tab:appendixA_case_study}
\setlength{\tabcolsep}{6pt}
\renewcommand{\arraystretch}{1}
\begin{tabularx}{0.98\textwidth}{@{}l X@{}}
\toprule
\textbf{Item} & \textbf{Content} \\
\midrule
\multicolumn{2}{@{}l@{}}{\textbf{Prompt definitions}} \\
\addlinespace[2pt]
Prompt A (accuracy = 0.852) &
Analyze the given statements about individuals telling the truth or lying, and determine the truthfulness of a specific person by iteratively applying logical deductions, stating assumptions before solving, and considering the chain of statements to arrive at a conclusion. \\
Prompt B (accuracy = 1.000) &
Analyze the given statements about individuals telling the truth or lying, and determine the truthfulness of the person in question by following the chain of statements and applying logical reasoning. For example, if we have the statements: `Fidel lies. Fletcher says Fidel tells the truth. Yoland says Fletcher lies. Raymond says Yoland lies. Leda says Raymond lies,' we can conclude that Leda tells the truth by tracing the statements: since Fidel lies, Fletcher's statement that Fidel tells the truth is false, meaning Yoland's statement that Fletcher lies is true, which in turn means Raymond's statement that Yoland lies is false, and thus Leda's statement that Raymond lies is true. Another example is: `Willian tells the truth. Phoebe says Willian lies. Alejandro says Phoebe lies. Lorine says Alejandro tells the truth. Christie says Lorine tells the truth,' where we can deduce Christie tells the truth by following the chain of statements in a similar manner. Apply this reasoning process to the given question to determine the truthfulness of the person in question. \\
\midrule
\multicolumn{2}{@{}l@{}}{\textbf{Case I: Both prompts output the \emph{same correct} answer $\rightarrow$ Reasoning-based judge}} \\
\addlinespace[2pt]
Question &
Kristian lies. Sherrie says Kristian lies. Delbert says Sherrie lies. Jerry says Delbert tells the truth. Shalonda says Jerry tells the truth. \textit{Does Shalonda tell the truth?} \\
Correct answer & No \\
Prompt A reasoning &
Since Kristian lies and Sherrie says Kristian lies, Sherrie tells the truth. Delbert says Sherrie lies, so Delbert lies, which means Jerry lies when saying Delbert tells the truth, and consequently, Shalonda lies when saying Jerry tells the truth. \\
Prompt A answer & No \\
Prompt B reasoning &
Since Kristian lies, Sherrie's statement that Kristian lies is true. This means Delbert's statement that Sherrie lies is false, so Jerry's statement that Delbert tells the truth is false, and thus Shalonda's statement that Jerry tells the truth is false. \\
Prompt B answer & No \\
Judge reasoning &
Prompt B's reasoning is more concise and directly addresses the truthfulness of each statement in a straightforward manner. It clearly states the implications of each person's statement without introducing unnecessary complexity, making it easier to follow and understand. The logical flow from Kristian's lie to Shalonda's falsehood is more streamlined in B, enhancing clarity and coherence. \\
\textbf{Judge decision} & \textbf{B} \\
\bottomrule
\end{tabularx}
\end{table*}

\clearpage
\begin{table*}[t]
\centering
\setlength{\tabcolsep}{6pt}
\renewcommand{\arraystretch}{1.18}
\begin{tabularx}{0.98\textwidth}{@{}l X@{}}
\toprule
% \multicolumn{2}{@{}l@{}}{\textbf{Case II: Prompts output \emph{different} answers (B correct, A incorrect) $\rightarrow$ Answer-based judge}} \\
\multicolumn{2}{@{}l@{}}{\textbf{Case II: Prompts output \emph{different} answers (B correct, A incorrect)}} \\
\multicolumn{2}{@{}l@{}}{\hspace*{4.3em}\textbf{$\rightarrow$ Answer-based judge}} \\
\addlinespace[2pt]
Question &
Vina tells the truth. Helene says Vina lies. Kandi says Helene tells the truth. Jamey says Kandi lies. Ka says Jamey lies. \textit{Does Ka tell the truth?} \\
Correct answer & No \\
Prompt A reasoning &
If Ka lies, then Jamey tells the truth, which means Kandi lies, and thus Helene lies, contradicting Vina's truthfulness. Therefore, Ka must tell the truth to maintain the consistency of the statements. \\
Prompt A answer & Yes \\
Prompt B reasoning &
Since Vina tells the truth, Helene's statement that Vina lies is false. This makes Kandi's statement that Helene tells the truth also false, which in turn makes Jamey's statement that Kandi lies true, and thus Ka's statement that Jamey lies is false, meaning Ka lies. \\
Prompt B answer & No \\
Judge reasoning &
The correct answer is determined by tracing the statements back to Vina's truthfulness. Response B correctly identifies the chain: Vina true $\rightarrow$ Helene false $\rightarrow$ Kandi false $\rightarrow$ Jamey true $\rightarrow$ Ka's statement about Jamey false (Ka lies). This matches the task requirement. \\
\textbf{Judge decision} & \textbf{B} \\
\bottomrule
\end{tabularx}
\end{table*}

\clearpage

\begin{examplequery}{BBH: Answer-based Preference Judge Template}
## Role ##
You are a specialized judge focused on evaluating answer correctness when two responses give different answers.
## Task ##
{question}
## Response from Prompt X ##
**Reasoning:** {reasoning_X}
**Answer:** {answer_X}
## Response from Prompt Y ##
**Reasoning:** {reasoning_Y}
**Answer:** {answer_Y}
## Your Task ##
The responses above give different answers: "{answer_X}" vs "{answer_Y}".
Your job is to determine which answer is more correct for the given task.
## Evaluation Criteria ##
Focus primarily on:
1. **Factual Accuracy** - Which answer better matches reality and task requirements?
2. **Task Alignment** - Which answer better fulfills the specific question asked?
## Output Format ##
{{
  "reasoning": "Your detailed justification explaining why prompt X or Y provided the more correct answer (~100 words).",
  "winner": "X or Y"
}}
\end{examplequery}

\begin{examplequery}{BBH: Reasoning-based Preference Judge Template}
## Role ##
You are a specialized judge focused on evaluating reasoning quality when two responses give the same answer.
## Task ##
{question}
## Response from Prompt X ##
**Reasoning:** {reasoning_X}
**Answer:** {answer_X}
## Response from Prompt Y ##
**Reasoning:** {reasoning_Y}
**Answer:** {answer_Y}
## Your Task ##
The responses above give the same answer: "{answer_X}".
Since both arrive at the same conclusion, your job is to determine which reasoning process is better.
## Evaluation Criteria ##
Focus primarily on:
1. **Logical Coherence** - Is the reasoning chain clear and well-structured?
2. **Completeness** - Does the reasoning address all key aspects of the problem?
3. **Clarity** - Is the reasoning easy to follow and understand?
4. **Accuracy** - Are the intermediate steps and assumptions correct?
## Output Format ##
{{
  "reasoning": "Your detailed justification explaining why prompt X or Y provided better reasoning (~100 words).",
  "winner": "X or Y"
}}
\end{examplequery}

\begin{examplequery}{MS-MARCO: Preference Judge Template}
## Role ##
You are a meticulous, impartial referee evaluating two competing answers to determine which better answers the given question based on the provided context.
## Query ##
{query}
## Context ##
{context}
## Answer from Prompt X ##
{answer_X}
## Answer from Prompt Y ##
{answer_Y}
## Evaluation Criteria ##
Compare both answers based on:
1. **Accuracy** - How factually correct is each answer based on the context?
2. **Completeness** - Does the answer address all aspects of the question?
3. **Relevance** - How well does the answer stay focused on answering the question?
4. **Clarity** - How clear and well-articulated is the answer?
## Output Format ##
{{
    "reasoning": "Your detailed justification explaining why answer X or Y is better (~100 words).",
    "winner": "X or Y"
}}
\end{examplequery}

\begin{examplequery}{MS-MARCO: final evaluations with ground-truth references}
"""
Begin your evaluation by carefully comparing the AI-generated answer with the reference solution. Identify any outputs that are not true, as well as omissions or deviations, and simulate human judgments in explaining how these impact the overall quality of the response. Ensure that your assessment is objective and consistent.
At the end of your evaluation, assign a score from 1 to 5 based on the following scale:

- 1: Very poor - does not meet the requirements or is significantly incorrect.
- 2: Poor - contains major errors or omissions.
- 3: Fair - adequate but with notable flaws.
- 4: Good - meets the requirements with minor errors.
- 5: Excellent - fully accurate and well-articulated.

[User Question]
{question}

[Reference Solution]
{ground_truth}

[AI-Generated Answer]
{prediction}

Your response should be a valid JSON string (no backticks) following this schema:
{{ "explanation": "{{Detailed reasoning based on the comparison}}" 
    "score": {{1-5}} 
}}
"""
\end{examplequery}

\begin{examplequery}{Template for generating initial prompts}
"""
You are an expert prompt-engineer. Your task is to generate **exactly 1** *high-quality* **system-level instruction** for the target reasoning task.

# Dataset Snapshot
Below is an *LM-written summary* of the unlabeled question pool:
{dataset_summary}

# Sample Inputs (do NOT answer them)
{questions}

# Prompt-Engineering Tip
{tip}

# Output Format (STRICT)
Return **exactly** 1 item in a JSON array, *and nothing else*:
[ "Your single high-quality instruction here" ]
"""
\end{examplequery}

\begin{examplequery}{Template for mutating on top-performing prompts}
"""
You are an expert prompt-engineer specializing in prompt optimization.
Your task is to generate 1 *diverse, high-quality* **mutation** of the currently **BEST PERFORMING** instruction for the target reasoning task.

# BEST PERFORMING Instruction (Current Champion)
{instructions}

# CRITICAL: Follow This Prompt-Engineering Tip
{tip}

# Output Format
Return **exactly** 1 mutated instruction in a JSON object, *and nothing else*:
{ "mutated_prompt": "Your mutated instruction here, following the tip." }
"""

\end{examplequery}

\begin{examplequery}{Prompt engineering tips for prompt generations}
# Mutation tips for top-performer-guided promp mutations
MUTATION_TIPS = {
    "expansion": "Keep the current champion instruction exactly as is, but expand on it by adding additional helpful guidance or clarifications. The result should be the original instruction plus new supplementary content.",
    "minimal": "Make very minimal changes to the current champion instruction. Keep it around the same length and modify only a few words through paraphrasing while preserving the core meaning.",
    "few_shot": "Add a few concrete examples to the current champion instruction to demonstrate the expected reasoning process or output format. Include 1-3 brief example cases that show how to apply the instruction.",
    "emphasis": "Adjust the tone, emphasis, or directional focus of the current champion instruction to create different reasoning patterns.",
}

# Initial instruction generation tips
INITIAL_INSTRUCTION_TIPS = {
    "framing": "Set the context for the task by framing it as a concrete creative scenario.",
    "simple": "Keep the instruction clear and concise.",
    "description": "Make sure your instruction is very informative and descriptive.",
    "persona": "Provide the LM with a creative persona that is relevant to the task.",
    "edge_cases": "List tricky cases the instruction must handle.",
    "assumptions": "Have the model state assumptions before solving.",
}
\end{examplequery}

\end{document}